\documentclass[twoside]{article}

\usepackage[accepted]{aistats2023}

\usepackage{hyperref}
\usepackage{url}
\usepackage[numbers]{natbib} 
\usepackage{multirow}
\usepackage{array}
\usepackage{amsmath, amssymb}
\usepackage{bbm}
\usepackage{caption}
\DeclareCaptionLabelFormat{AppendixTables}{Table A.#2}
\usepackage{subcaption}

\usepackage{mathtools} 
\usepackage{booktabs} 
\usepackage{tikz} 
\usepackage{wrapfig,lipsum,booktabs}
\usepackage{enumitem}
\newcommand{\beginsupplement}{%
        \setcounter{table}{0}
        \renewcommand{\thetable}{A\arabic{table}}%
        \setcounter{figure}{0}
        \renewcommand{\thefigure}{A\arabic{figure}}}%

\definecolor{darkpastelgreen}{rgb}{0.01, 0.75, 0.24}

\let\subparagraph\paragraph
\let\subparagraph\paragraph
\usepackage[compact]{titlesec}

\titlespacing{\section}{0pt}{1ex}{1ex}
\titlespacing{\subsection}{0pt}{1ex}{0.5ex}
\titlespacing{\subsubsection}{0pt}{0.5ex}{0.5ex}

\DeclareMathSizes{9}{8.2}{5}{3}
\begin{document}
\runningauthor{Samaddar, Madireddy, Balaprakash, Maiti, de los Campos, Fischer}
\twocolumn[

\aistatstitle{Sparsity-Inducing Categorical Prior Improves Robustness of the Information Bottleneck}

\aistatsauthor{ Anirban Samaddar \And Sandeep Madireddy \And  Prasanna Balaprakash }

\aistatsaddress{ Michigan State University \And  Argonne National Laboratory \And Argonne National Laboratory }

\aistatsauthor{Tapabrata Maiti \And Gustavo de los Campos \And Ian Fischer}

\aistatsaddress{ Michigan State University \And  Michigan State University \And Google Research } ]

\begin{abstract}

The information bottleneck framework provides a systematic approach to
learning representations that compress nuisance information in the input 
and extract semantically meaningful information about predictions. 
However, the choice of a prior distribution that fixes the dimensionality
across all the data can restrict the flexibility of this approach for 
learning robust representations. We present a novel sparsity-inducing
spike-slab categorical prior that uses sparsity as a mechanism to provide 
the flexibility that allows each data point to learn its own dimension distribution. 
In addition, it provides a mechanism for learning a joint distribution of the 
latent variable and the sparsity hence can account for the complete uncertainty 
in the latent space. Through a series of experiments using in-distribution and
out-of-distribution learning scenarios on the MNIST, CIFAR-10 and ImageNet data, we 
show that the proposed approach improves accuracy and
robustness compared to traditional fixed-dimensional priors, as well as other
sparsity induction mechanisms for latent variable models proposed in the literature.

\end{abstract}

\section{Introduction}\label{sec:intro}

Information bottleneck (IB) (\cite{tishby2000information}) is a
deep latent variable model that poses representation compression as a constrained 
optimization problem to find representations $Z$ that are maximally informative 
about the outputs $Y$ while being maximally compressive about the inputs $X$, 
using a loss function expressed using a mutual information (MI) metric 
and a Lagrangian formulation of the constrained optimization: 
$\mathcal{L}_{IB} = \operatorname{MI}(X;Z) - \beta \operatorname{MI}(Z;Y)$.
Here, $\operatorname{MI}(X;Z) $ is the MI that reflects how much the 
representation compresses $X$, and $\operatorname{MI}(Z;Y)$ reflects how much information 
the representation has kept from $Y$.

It is standard practice to use parametric priors, such as a mean-field Gaussian 
prior for the latent variable $Z$, as seen with most latent-variable models in the 
literature (\cite{tomczakdeep}). In general, however, a major limitation of these priors is 
the requirement to preselect a latent space complexity for all data, which can be 
very restrictive and lead to models that are less robust. Sparsity, when used as a 
mechanism to choose the complexity of the model in a flexible and data-driven fashion, 
has the potential to improve the robustness of machine learning systems without loss of accuracy, especially when dealing with high-dimensional data (\cite{ahuja2021invariance}). 

\begin{figure}
    \centering
    \includegraphics[width = 0.45\textwidth]{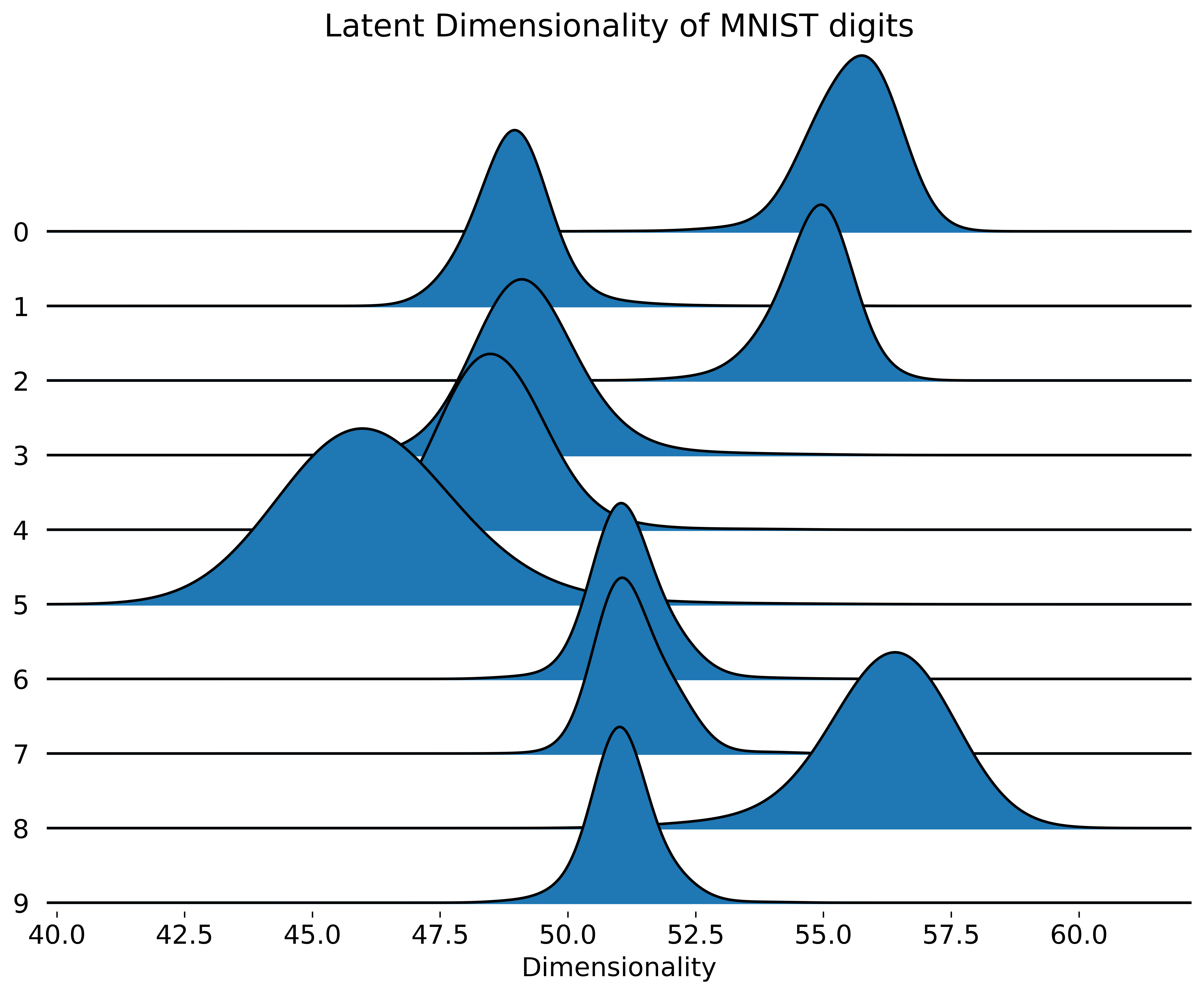}
    \caption{Plot of latent dimension distribution learned by \texttt{SparC-IB} aggregated for each of the MNIST data classes. In the vertical axis, we have 10 digits classes and in the horizontal axis we have their distribution of posterior modes of latent dimension aggregated across the testset data points. It shows our \texttt{SparC-IB} prior provides flexibility to learn the data-specific latent dimension distribution, in contrast to fixing to a single value.
    \vspace{-0.05in}}
\label{fig:Dimension_vs_digit_MNIST}
\end{figure}
Sparsity has been considered in the context of latent variable models in a handful of works. In linear latent variable modeling, \cite{lan2014sparse} proposes a sparse factor model in the context of learning analytics, \cite{chun2010sparse} proposes a sparse partial least squares (sPLS) method to resolve inconsistency issue that arises in standard PLS in a high-dimensional setting, and \cite{yang2019survey} reviews advances in sparse canonical correlation analysis. Most of the work in sparse linear latent variable modeling fixes the latent dimensionality or treats it as a hyperparameter. In nonlinear latent variable modeling, sparsity was proposed mostly in the 
context of unsupervised learning. Notable works include sparse Dirichlet variational autoencoder (sDVAE) (\cite{burkhardt2019decoupling}), epitome VAE 
(eVAE) (\cite{yeung2017tackling}), variational sparse coding (VSC) (\cite{tonolini2020variational}), and InteL-VAE (\cite{miao2022incorporating}). 
Most of these approaches do not take into account the uncertainty involved in introducing sparsity and treat this as a deterministic selection problem. Approaches such as the Indian buffet process VAE \cite{singh2017structured} relax this and allow learning a distribution over the selection parameters that induce sparsity, but only allow global sparsity. 
Ignoring uncertainty in selection and the flexibility to learn a local data-driven dimensionality of the latent space for each data point can lead to a loss of robustness in inference and prediction. 


We also note that the aforementioned approaches have been proposed for unsupervised learning, and 
we are interested in the supervised learning scenario introduced with the IB approach, which poses 
a different set of challenges because the sparsity has to accommodate accurate prediction $Y$. 
Only one recent work (\cite{kim2021drop}) that we know of has considered sparsity in the latent variables
of the IB model, but here the latent variable is assumed to be deterministic, and the sparsity applied
indirectly by weighting each dimension using a Bernoulli distribution, where zero weight is equivalent to
sparsification.

\noindent To that end, we make the following contributions:
\begin{enumerate}[leftmargin=*,topsep=0pt,itemsep=-1ex,partopsep=1ex,parsep=1ex]
\item We introduce a novel sparsity-inducing Bayesian spike-slab prior for supervised learning in the IB framework, where, the sparsity in the latent variables of the IB model is modeled stochastically through a categorical distribution, and thus the joint distribution of latent variable and sparsity 
is learned with this categorical prior IB (\texttt{SparC-IB}) model through a Bayesian inference.
\item We derive variational lower bounds for efficient inference of the proposed \texttt{SparC-IB} model 
\item Using in-distribution and out-of-distribution experiments with MNIST, CIFAR-10, and ImageNet data, we show improvement in accuracy and robustness with \texttt{SparC-IB} compared to vanilla VIB models and other sparsity-inducing strategies.
\item Through extensive analysis of the latent space, we show that learning the joint distribution provides the flexibility to systematically learn parsimonious data-specific (local) latent dimension complexity (as shown in \hyperref[fig:Dimension_vs_digit_MNIST]{Fig. 1}), 
enabling them to learn robust representations.
\end{enumerate}


\section{Related Works}
Previous works in the literature on latent-variable models has looked at different sparsity-inducing mechanisms in the latent space in supervised and unsupervised settings. \cite{burkhardt2019decoupling} propose a sparse Dirichlet variational autoencoder (sDVAE) that assumes a degenerate Dirichlet distribution on the latent variable. They introduce sparsity in the concentration parameter of the Dirichlet distribution through a deterministic neural network. Epitome VAE (eVAE) by \cite{yeung2017tackling} imposes sparsity through epitomes that select adjacent coordinates of the mean-field Gaussian latent vector and mask the others before feeding to the decoder. The authors propose training a deterministic epitome selector to select from all possible epitomes. Similar to eVAE, the variational sparse coding (VSC) proposed in \cite{tonolini2020variational} introduces sparsity through a deterministic classifier. 

\begin{table}[h!]
    \centering
    \resizebox{0.7\columnwidth}{!}{
    \begin{tabular}{|c|c|c|}
      \hline
      Approach & Latent & Sparsity \\
               & Variable & (global/local)\\ \hline
       sDVAE  & S  & D (L) \\ \hline
       eVAE  & S & D (L)\\ \hline
       VSC  & S & D (L)\\ \hline
       InteL-VAE  & S & D (L)\\ \hline
       Drop-B  & D & S (G)\\ \hline
       IBP  & S & S (G)\\ \hline
       \texttt{SparC-IB} & S & S (L)\\ \hline
    \end{tabular}
    }
    \caption{Latent-variable models with different sparsity induction strategies, where D=Deterministic and S=Stochastic. The type of induced sparsity is in parentheses, where G is global sparsity and L the local sparsity.
    \vspace{-0.05in}}
    \label{tab:my_label}
\end{table}
In this case, the classifier outputs an index from a set of pseudo inputs that define the prior for the latent variable. In a more recent work, InteL-VAE (\cite{miao2022incorporating})  introduces sparsity via a dimension selector (DS) network on top of the Gaussian encoder layer in standard VAEs. The output of DS is multiplied with the Gaussian encoder to induce sparsity and then is fed to the decoder. InteL-VAE has empirically shown an improvement over VSC in unsupervised learning tasks, such as image generation.
The Indian buffet process (IBP) (\cite{singh2017structured}) learns a distribution on infinite-dimensional sparse matrices where each element of the matrix is an independent Bernoulli random variable, where the probabilities are global and come from a Beta distribution. Therefore, the sparsity induced by IBP is global and can make the coordinates of the latent variable zero for all data points.

In the IB literature we find the aspect of sparsity in the latent space seldom explored. To the best of our knowledge, Drop-Bottleneck (Drop-B) by \cite{kim2021drop} is the only work that attempts this problem. In Drop-B, a neural network extracts features from the data; then, a Bernoulli random variable stochastically drops certain features before passing them to the decoder. The probabilities of the Bernoulli distribution, similar to IBP, are assumed as global parameters and are trained with other parameters of the model. 

In this paper, with \texttt{SparC-IB}, we model stochasticity in both latent variables and sparsity, and we relax the global sparsity assumption by learning the distribution of (local) sparsity for each data point. In this regard, we differ from other latent-variable models.
In \hyperref[tab:my_label]{Table 1} we summarize these different approaches by the types, stochastic (S) or deterministic (D), of the latent variable and the sparsity. Furthermore, we characterize the sparsity induced by each method by whether they impose  global (G) or local (L) sparsity. The table shows that very few works incorporate stochasticity in both latent variable and sparsity-inducing mechanism. 

\section{Information Bottleneck with Sparsity-Inducing Categorical Prior}
\subsection{Information Bottleneck: Preliminaries}
Taking into account a joint distribution $P(X,Y)$ of the input variable $X$  and the corresponding 
target variable $Y$, the information bottleneck principle aims to find a (low-dimensional) latent encoding $Z$ 
by maximizing prediction accuracy, formulated in terms of mutual information $\operatorname{MI}(Z;Y)$, 
given a constraint on compressing the latent encoding, formulated in terms of mutual 
information $\operatorname{MI}(X;Z)$. 
This can be cast as a constrained optimization problem:
\begin{equation}
\begin{array}{ll}
\label{eq:constrained_IB}
&\max_{Z} \mathrm{MI}(Z; Y) \\
&\textsc{s.t.} \quad \mathrm{MI}(X; Z)\leq C,
\end{array}
\end{equation}
where $C$ can be interpreted as the compression rate or the minimum number of bits needed to 
describe the input data. Mutual information is obtained through a multidimensional integral that depends on the joint distribution and the marginal distribution of random variables given by 
    $
    \int_{\mathcal Z} \int_{\mathcal X}
        {P_{\theta}(x,z) \log{ \left(\frac{P_{\theta}(x,z)}{P(x)\,P(z)} \right) }
    } \; dx \,dz,$
where $P_\theta(x,z) := P_\theta(z|x)P(x)$; a similar expression for $\operatorname{MI}(Z;Y)$ needs 
$P_\theta(y,z) := \int P_\theta(z|x)P(x,y) \mathrm{d} x \,.$ The integral presented to calculate $\operatorname{MI}$ is generally computationally intractable for large data. 

Thus, in practice, the Lagrangian relaxation of the constrained optimization problem is 
adopted~\cite{tishby2015deep}:
\begin{equation}
  \mathcal{L}_{IB}(Z) = \mathrm{MI}(Z; Y) - \beta \ \mathrm{MI}(X; Z)
  \label{eq:IB_lagrangian}
\end{equation}
where $\beta$ is a Lagrange multiplier that enforces the constraint $\mathrm{MI}(X; Z) \leq C$ 
such that a latent encoding $Z$ is desired that is maximally expressive about $Y$ while being maximally 
compressive about $X$. In other words, $\mathrm{MI}(X;Z) $ is the mutual information 
that reflects how much the representation ($Z$) compresses $X$,  
and $\mathrm{MI}(Z;Y)$ reflects how much information the representation has been kept from Y.
Several approaches have been proposed in the literature to approximate mutual information 
$\mathrm{MI}(X;Z)$, ranging from parametric bounds defined by variational lower 
bounds (\cite{alemi2016deep}) to non-parametric bounds (based on kernel density estimate) (\cite{Kolchinsky19}) and adversarial f-divergences \cite{zhai2021adversarial}. 
In this research, we focus primarily on the variational lower bounds-based approximation. Furthermore, we take a square transformation of the term $\mathrm{MI}(X;Z)$ following \cite{Kolchinsky19}. Taking a convex transformation of the compression term makes the solution of the IB Lagrangian identifiable w.r.t. $\beta$ (\cite{Galvez2020}). However, for the sake of clarity, we drop this transformation from the derivation of the loss function. From the convexity property, all derivations with standard IB loss
carry over to the loss function with this transformation.


\noindent \paragraph{Role of Prior Distribution and the Stochasticity of the Latent Variable}
In the IB formulation presented above, the latent variable is assumed to be stochastic and, hence, 
the posterior distribution of it is learned by using Bayesian inference. In \cite{alemi2016deep}, 
the variational lower bound of $\mathcal{L}_{IB}(Z)$ is given as
{\small
\begin{align}
  \mathcal{L}_{VIB}(Z) &=  \mathbb{E}_{X,Y} \bigg[ \mathbb{E}_{Z|X} \log q(Y|Z) - \beta  \  \operatorname{D}_{\operatorname{KL}}(q(Z|X) || q(Z)) \bigg] &
  \label{eq:VIB_loss}
\end{align}
}
In equation \hyperref[eq:VIB_loss]{(3)}, the prior $q(z)$ serves as a penalty to the variational encoder $q(z|x)$ and $q(y|z)$ is the decoder that is the variational approximation to $p(y|z)$. \cite{alemi2016deep} choose the encoder family and the prior to be a fixed $K$-dimensional isotropic multivariate Gaussian distribution ($N(0_K,I_{K\times K})$). This choice is motivated by the simplicity and efficiency of inference using differentiable reparameterization of the Gaussian random variable $Z$ in terms of its mean and sigma. \emph{For complex datasets, however, this can be restrictive since the same dimensionality is imposed on all the data and hence can prohibit the latent space from learning robust features} (\cite{miao2022incorporating}).

We describe a new family of sparsity-inducing priors 
that allow stochastic exploration of different dimensional latent spaces. For the proposed variational family, we derive the variational lower bound that consists of discrete and continuous variables, and show that simple reparameterization steps can be taken to draw samples efficiently from the encoder 
for inference and prediction. 

\subsection{Sparsity-Inducing Categorical Prior}
\label{categ}
\begin{figure}[t!]
    \centering
    \includegraphics[width = 0.45\textwidth]{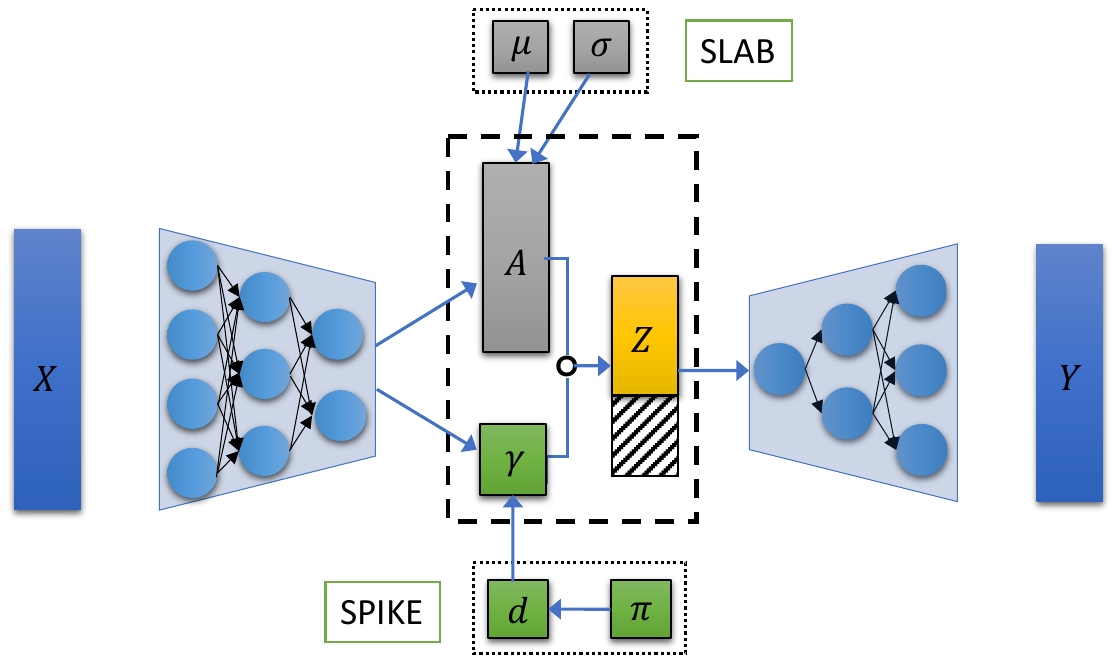}
    \caption{Schematic of the categorical prior information bottleneck (\texttt{SparC-IB}) model. The inputs are passed through the encoder layer to estimate the feature allocation vector $A$, while simultaneously using a categorical prior to sample the number of active dimensions $d$ of $A$ for  given data that selects the complexity of latent variable $Z$. The obtained $Z$ is then fed to the decoder to predict the supervised learning responses.} 
    \label{BBP_Schematic}
\end{figure}

A key aspect of the IB model and latent variable models in general is the dimension of $Z$. Fixing a very low dimension of $Z$ can impose a limitation on the capacity of the model, while a very large $Z$ can lead to learning a lot more nuisance information that can be detrimental for model robustness and generalizability.  We formulate a data-driven approach to learn the distribution of dimensionality $Z$ through the design of the sparsity-inducing prior.

Let $\{X_n,Y_n\}_{n=1}^{N}$ be $N$ data points with $X_n \in \mathbb{R}^{1 \times p}$ and let the latent variable $Z_n = (Z_{n,1},Z_{n,2},...,Z_{n,K})$, where $K$ is the latent dimensionality. The idea is that we will fix $K$ to be large a priori and make the prior assumption that the $k < K$ columns of $Z_n$ are zero and therefore do not contribute to the prediction of $Y_n$. Therefore, the prior distribution of $Z_n$ can be specified as follows.
\begin{flalign}
Z_{n,k}|\gamma_{n,k} \ &\sim \ (1-\gamma_{n,k}) \mathbbm{1}(Z_{n,k}=0) + \gamma_{n,k} \text{N}(\mu_{n,k}, \sigma_{n,k}^2) & \nonumber \\
& \text{OR} & \nonumber \\
Z_n &= A_n \circ \gamma_{n};
A_n \sim \mathcal{N}(\mu_n, \Sigma_n) & \\
\gamma_{n,k}|d_n \ &= \ \mathbbm{1}(k \leqslant d_n) ; 
d_n \sim \ \text{Categorical}(\pi_n)
\label{prior}
\end{flalign}
This prior is a variation of the spike-slab family (\cite{george1993variable}) where the sparsity-inducing parameters follow a categorical distribution. The latent variable $Z_n$ is an element-wise product between the feature allocation vector $A_n$ and a sparsity-inducing vector $\gamma_n$. The $K$ dimensional latent $A_n$ is assumed to follow a $K$ dimensional Gaussian distribution with mean vector $\mu_n$ and diagonal covariance matrix $\Sigma_n$ with the variance vector $\sigma^2_n$. Note that $\gamma_n$ is a vector whose first $d_n$ coordinates are 1s and the rest are 0s, and hence $A_n \circ \gamma_{n}$, with the result that the first $d_n$ coordinates of $Z_n$ are nonzero and the rest are 0. Therefore, as we sample $d_n$, we consider the different dimensionality of the latent space; $d_n$ follows a categorical distribution over the $K$ categories with $\pi_n$ as the vector of categorical probabilities whose elements $\pi_{n,k}$ denote the prior probability of the $k$th dimension for the data point $n$. In \hyperref[BBP_Schematic]{Fig. 2} we present the flow of the model from the input $X$ to the output $Y$ through the latent encoding layer.

\subsection{Derivation of the Variational Bound}
The variable $d_n$ augments $Z_n$ for latent space characterization. With a Markov chain assumption, $Y \leftrightarrow X \leftrightarrow Z$, the latent variable distribution factorization as $p(Z,d) = \prod_{n=1}^{N} p(Z_n|d_n) p(d_n)$, and the same family with the same factorization of the variational posterior $q(Z_n,d_n|X)$ as the prior, we derive the variational lower bound of the IB Lagrangian as follows. In the following derivation for simplicity, we use $X$ interchangeably for both the random variable $\in \mathbb{R}^{1 \times p}$ and the sample covariate matrix $\in \mathbb{R}^{N \times p}$.  
{\normalsize
\begin{flalign}
&\mathcal{L}_{IB}(Z) = \mathrm{MI}(Z; Y) - \beta \ \mathrm{MI}(X; Z) & \nonumber\\ 
&\geqslant \mathcal{L}_{VIB}(Z) &  \nonumber \\
&= \mathbb{E}_{X,Y} \bigg[ \mathbb{E}_{Z|X} \log q(Y|Z) - \beta \  \operatorname{D}_{\operatorname{KL}}(q(Z|X) || q(Z)) \bigg]  \ \nonumber & \\
& \hspace{4cm} \text{ (from \hyperref[eq:VIB_loss]{(3)})} & \nonumber \\
&= \mathbb{E}_{X,Y} \bigg[ \mathbb{E}_{Z|X} \log q(Y|Z) - \beta \  \operatorname{D}_{\operatorname{KL}}(q(Z,d|X) || q(Z,d)) \bigg] & \nonumber \\
&= \mathcal{L}_{\texttt{SparC-IB}}(Z,d) & \nonumber
\end{flalign}
}%
Note that we have introduced the random variable $d$ along with $Z$ in the density function of the second KL-term. This is because the KL-divergence between $q(Z|X)$ and $q(Z)$ is intractable. The equality of the loss function is valid since $d_n = \sum_{k=1}^{K} \mathbbm{1}(Z_{n,k} \neq 0)$ is a deterministic function of $Z_n$, and we can write the density $q(z,d) = q(z)q(d|z) = q(z)\delta_{d_z}(d)$, where $\delta_{c}(d)$ is the Dirac delta function at $c$. Therefore, we have the following.
\begin{flalign}
&\operatorname{D}_{\operatorname{KL}}(q(Z,d|X)||q(Z,d)) 
= \mathbb{E}_{Z,d|X} \log \frac{q(Z|X)\delta_{d_Z(d)}}{q(Z)\delta_{d_Z(d)}} \nonumber \\
&= \mathbb{E}_{Z|X} \mathbb{E}_{d|Z} \log \frac{q(Z|X)\delta_{d_Z}(d)}{q(Z)\delta_{d_Z}(d)}  
= \mathbb{E}_{Z|X} \log \frac{q(Z|X)}{q(Z)} \nonumber  \\ 
&= \operatorname{D}_{\operatorname{KL}}(q(Z|X)||q(Z))  \nonumber
\end{flalign}
Note that $\delta_{d_Z}(d) = 1$, given $Z$, almost everywhere since $\delta_{d_Z}(d) = 0 \iff d \neq d_z$ has measure 0 under $q(d|z) = \delta_{d_z}(d)$. We now replace $\mathbb{E}_{X,Y}$ with the empirical version.
\begin{flalign}
\mathcal{L}_{\texttt{SparC-IB}}(Z,d) &\hat{=} \frac{1}{N} \sum_{n=1}^{N} \bigg[ \mathbb{E}_{Z_n|X} [\log q(Y_n|Z_n)] \nonumber & \\ & \hspace{2cm} - \beta \  \operatorname{D}_{\operatorname{KL}}(q(Z_n,d_n|X) || q(Z_n,d_n)) \bigg] & \nonumber \\
& = \frac{1}{N} \sum_{n=1}^{N} \bigg[ \mathbb{E}_{d_n|X} \mathbb{E}_{Z_n|X,d_n} [\log q(Y_n|Z_n)] \nonumber & \\ & 
\hspace{1cm} - \beta \ \big[ \mathbb{E}_{d_n|X}[ \operatorname{D}_{\operatorname{KL}}(q(Z_n|X,d_n) || q(Z_n|d_n))]  & \nonumber \\
&\hspace{2cm} + \operatorname{D}_{\operatorname{KL}}(q(d_n|X)||q(d_n)) \big] \bigg] & \nonumber
\end{flalign}
\noindent We analyze the three terms in the above decomposition as follows.
\begin{flalign}
(i) \ & \mathbb{E}_{d_n|X} \mathbb{E}_{Z_n|X,d_n} [\log q(Y_n|Z_n)] & \nonumber \\ 
&= \sum_{k=1}^{K} \mathbb{E}_{Z_n|X,d_n = k} [\log q(Y_n|Z_n,d_n=k)] \pi_{n,k}(X)  & \nonumber
\end{flalign}
This term is a weighted average of the negative cross-entropy losses from models with increasing dimension of latent space, where the weights are the posterior probabilities of the dimension encoder. Therefore, \emph{maximizing this term implies putting large weights on the dimensions of the latent space where log-likelihood is high}. During training, this term can be computed using the Monte Carlo approximation, that is,  $ (i) \hat{=} \frac{1}{J} \sum_{j=1}^{J} \log q(Y_n|Z_n = Z_n^{(j)}, d_n = d_n^{(j)})$, where we draw $J$ randomly drawn samples from $q(Z_n,d_n|X)$. In our experiments, we fixed $J=10$ everywhere during training.
\begin{flalign}
(ii) \ & \mathbb{E}_{d_n|X}[ \operatorname{D}_{\operatorname{KL}}(q(Z_n|X,d_n) || q(Z_n|d_n))] \nonumber \\
&= \sum_{k=1}^{K} \operatorname{D}_{\operatorname{KL}}(q(Z_n|X,d_n=k) || q(Z_n|d_n=k)) \pi_{n,k}(X)  \nonumber &
\end{flalign}
Note that $q(z_n|d_n) = q(z_n|\gamma_n) = \mathcal{N}(z_n;\Tilde{\mu}_n, \Tilde{\Sigma}_n)$, where $\Tilde{\mu}_n = \mu_n \circ \gamma_n$ and $\Tilde{\Sigma}_n$ are diagonal with the entries $\Tilde{\sigma}^2_n = \sigma^2_n \circ \gamma_n$. When $k < K$, this density does not exist w.r.t. the Lebesgue measure in $\mathbb{R}^K$. However, we can still define a density w.r.t. the Lebesgue measure restricted to $\mathbb{R}^k$ (see Chapter 8 in \cite{rao1973linear}), and it is the $k$-dimensional multivariate normal density with mean $\mu_{n,-\overline{K-k}}= (\mu_{n,1},...,\mu_{n,k})'$ and diagonal covariance matrix $\Sigma_{n,-\overline{K-k}}$ with diagonal entries $\sigma^2_{n,-\overline{K-k}} = (\sigma_{1,n}^2,...,\sigma_{n,k}^2)'$. Denoting $\mu_{n,-0} = \mu_n$, we have the following.
\begin{flalign}
(ii) term \ &=  \sum_{k=1}^{K} \operatorname{D}_{\operatorname{KL}}\bigg(\mathcal{N}(\mu_{n,-\overline{K-k}}(X),\Sigma_{n,-\overline{K-k}}(X)) \nonumber & \\
& \hspace{2cm} || \mathcal{N}(\mu_{n,-\overline{K-k}},\Sigma_{n,-\overline{K-k}})\bigg) \pi_{n,k}(X)  & \nonumber \\
&=\sum_{k=1}^{K} \sum_{\ell=1}^{k} \operatorname{D}_{\operatorname{KL}} \bigg(\mathcal{N}(\mu_{n,\ell}(X),\sigma^2_{n,\ell}(X)) \nonumber & \\
& \hspace{2cm} || \mathcal{N}(\mu_{n,\ell},\sigma^2_{n,\ell}) \bigg) \pi_{n,k}(X) \ & \nonumber \\
&= \frac{1}{2}  \sum_{k=1}^{K} \sum_{\ell=1}^{k} [\sigma_{n,\ell}^2(X) - 1 - \log(\sigma_{n,\ell}^2(X)) \nonumber & \\
& \hspace{2cm} + \mu_{n,\ell}^2(X)]\pi_{n,k}(X) \ & \nonumber
\end{flalign}
The second-last equality is due to the fact that the KL-divergence of multivariate Gaussian densities whose covariances are diagonal can be written as a sum of coordinate-wise KL-divergences. Since KL-divergence is always non-negative, \emph{minimizing the above expression implies putting more probability to the smaller-dimensional latent space models since the second summation term is expected to grow with dimension $k$}.
\begin{flalign}
(iii) \ \operatorname{D}_{\operatorname{KL}}(q(d_n|X)||q(d_n)) &= \sum_{k=1}^{K} \log \frac{\pi_{n,k}(X)}{\pi_{n,k}} \pi_{n,k}(X)  & \nonumber
\end{flalign}
\emph{Minimizing this term forces the learned probabilities to be close to the prior}. Note that we are learning these probabilities for each data point (since $\pi_{n,k}(X)$ is indexed by $n$). In this respect, we differ from most of the stochastic sparsity-inducing approaches, such as Drop-B (\cite{kim2021drop}) and IBP (\cite{singh2017structured}). In these approaches, the sparsity is induced from a probability distribution with global parameterization and is not learned for each data point.  

\noindent \paragraph{Modeling Choices for the \texttt{SparC-IB} Components:}
Since $Z_n = A_n \circ \gamma_n$, we are required to fix the priors for $(A_n,\gamma_n)$ or $(A_n, d_n)$. We chose $K$-dimensional spherical Gaussian $\mathcal{N}(0,I_K)$ as the prior for the latent variable $A_n$. For $d_n$, we assume that the $k$th categorical probability comes from the compound distribution of a beta-binomial model, also known as the {\it Polya urn} model~\cite{mahmoud2008polya}. Therefore,
\begin{equation}
    \label{compound}
    \pi_n = \mathbb{P}(d_n = k) = {K-1 \choose k-1} \frac{\text{B}(a_n+k-1,b_n+K-k)}{\text{B}(a_n,b_n)}.
\end{equation}
For simplicity, we set the prior value to be constant across the data points, that  is,  $(a_n,b_n) = (a,b)$. The key advantage of this choice is that we can write the probability as a differentiable function of the two shape parameters $(a_n,b_n)$. Therefore, we can assume the same categorical distribution for the encoder; and instead of learning $K$ probabilities $\pi_{n,k}(X)$ we can learn $(a_n(X),b_n(X))$, which significantly reduces the dimensionality of the parameter space. However, learning $\pi_{n,k}(X)$ provides more flexibility because the shape of the distribution is not constrained, while learning $(a_n(X),b_n(X))$ constrains $\pi_{n,k}(X)$ to follow according to the shape of the compound distribution, which depends on $a_n(X)$ and $b_n(X)$. In our experiments, we tested both approaches and found that learning $(a_n(X),b_n(X))$ produces better results.

\section{Experimental Results}
\label{Experiments}






\begin{figure*}[h!]
	\centering
	\begin{subfigure}{0.23\textwidth}
	\centering
	\includegraphics[width=\linewidth]{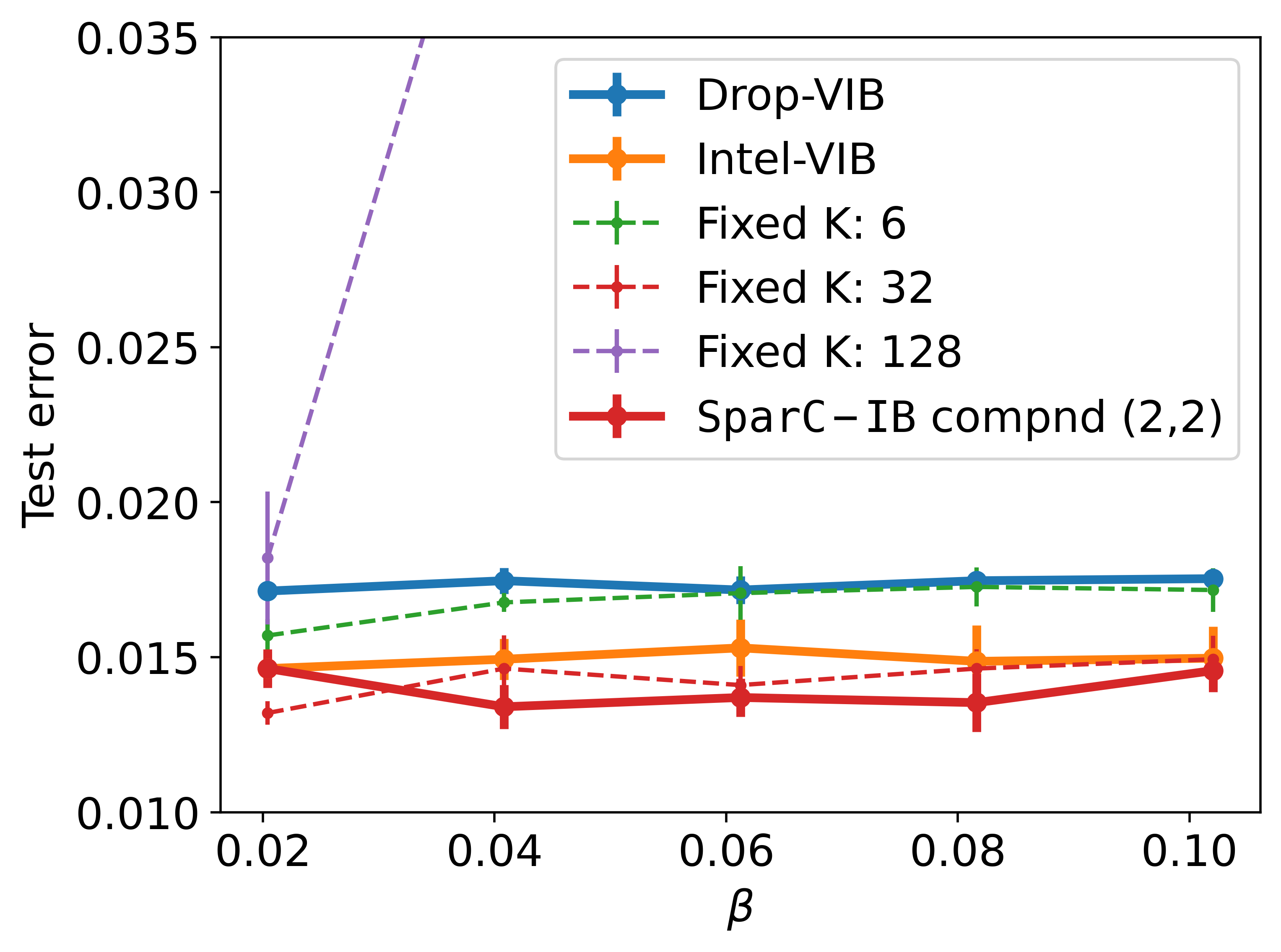}
	\caption{Test error vs. $\beta$ }
	\end{subfigure}
        \hfill
	\begin{subfigure}{0.23\textwidth}
	\centering
	\includegraphics[width=\linewidth]{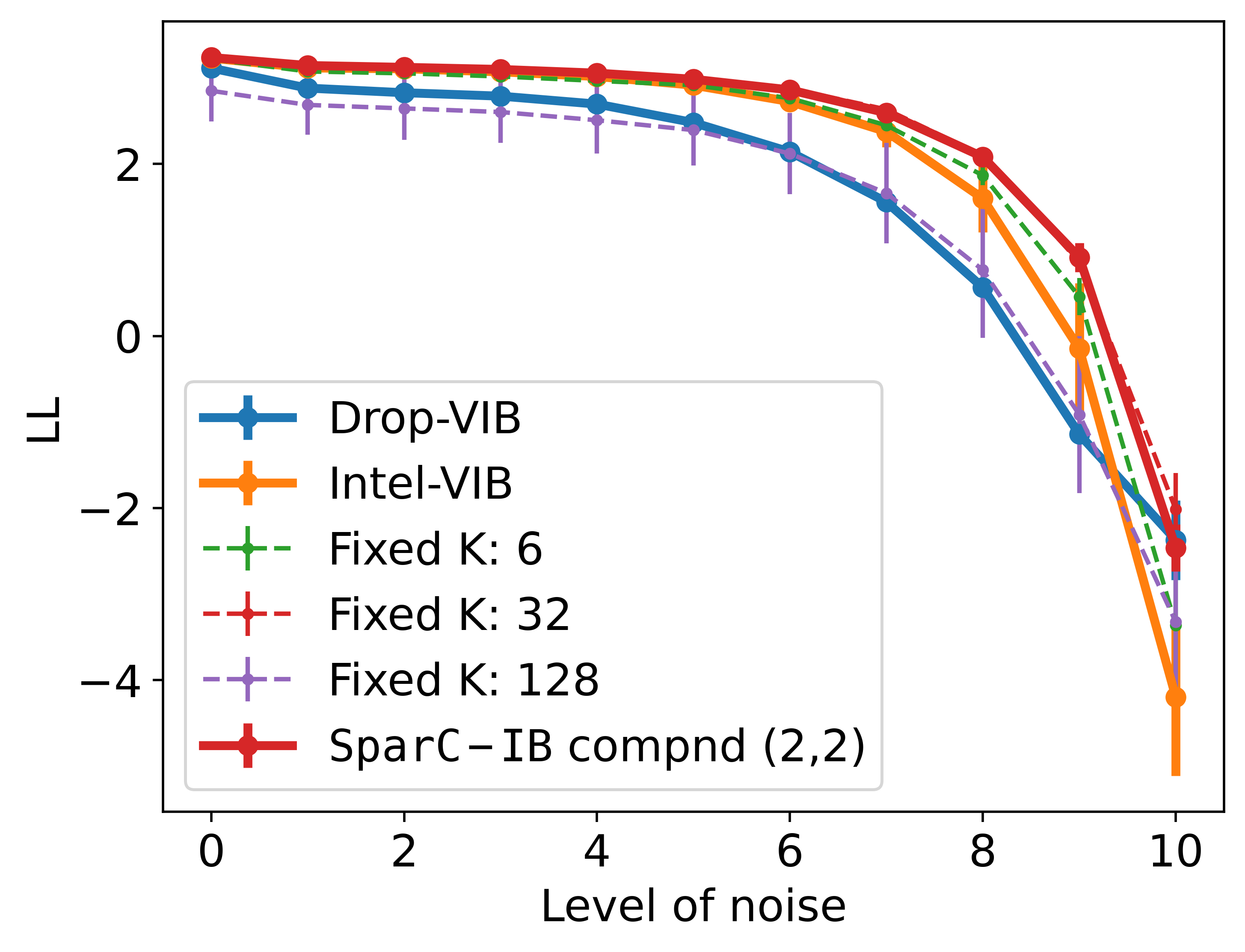}
	\caption{Log likelihood vs. noise}
	\end{subfigure}
        \hfill
	\begin{subfigure}{0.23\textwidth}
	\centering
	\includegraphics[width=\linewidth]{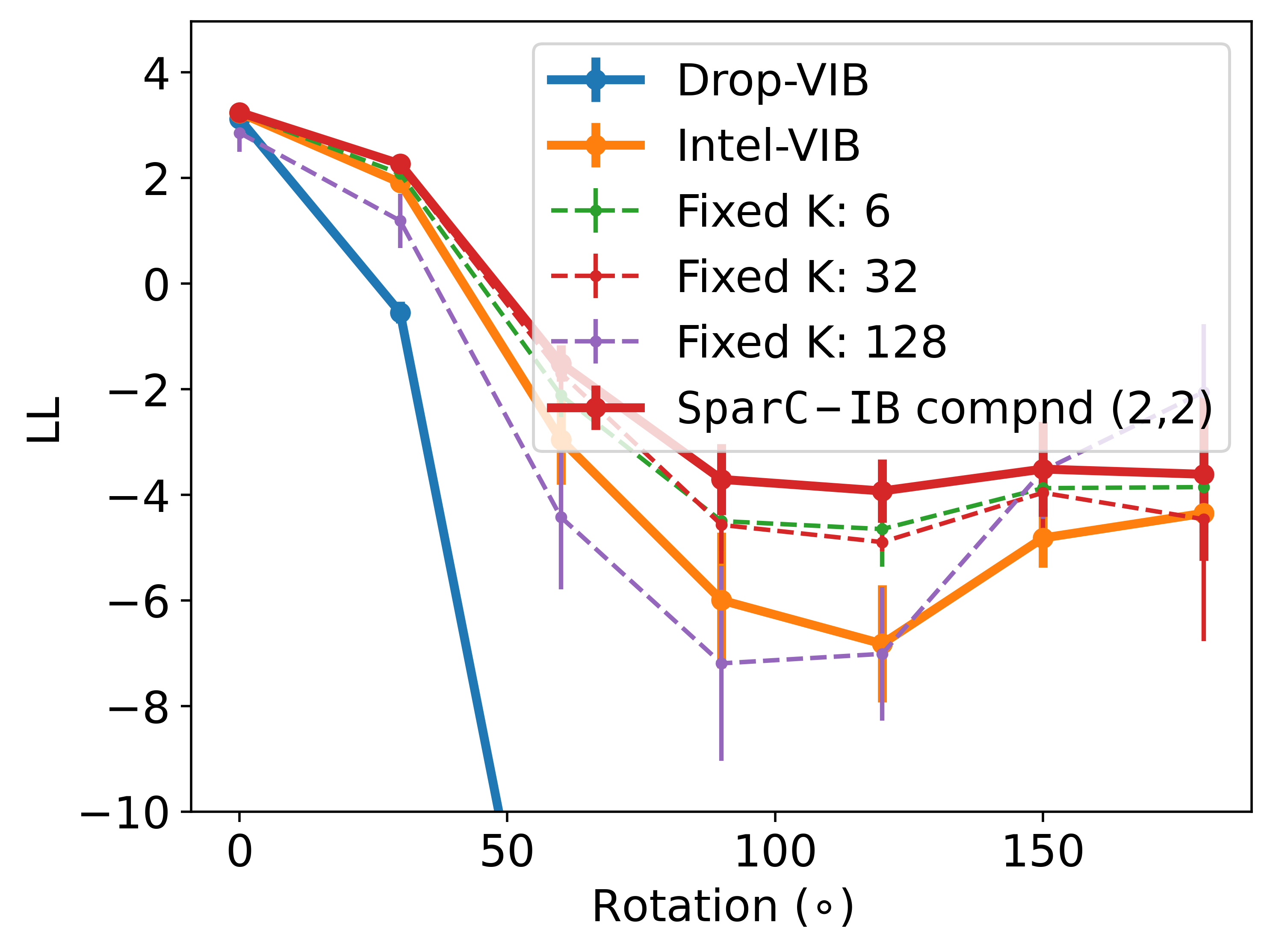}
	\caption{Log likelihood vs. rotation}
	\end{subfigure}
        \hfill
	\begin{subfigure}{0.23\textwidth}
	\centering
	\includegraphics[width=\linewidth]{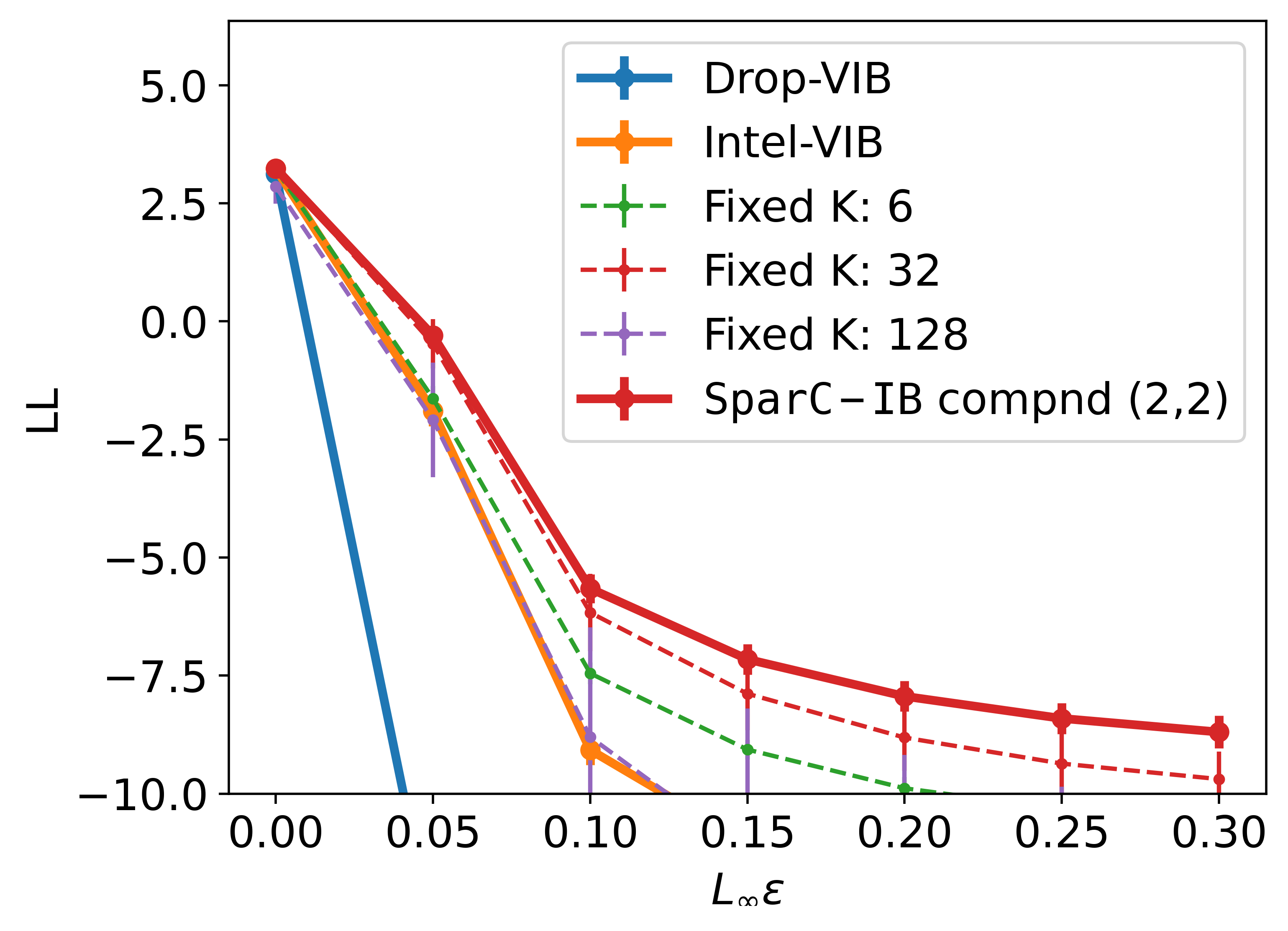}
	\caption{Log likelihood vs. attack}
	\end{subfigure}
        \vspace{-0.5\baselineskip}
 	\caption{In- and out-of- distribution performance on MNIST. \vspace{-0.15in}}
	\label{fig:Perf_MNIST}
\end{figure*}

\begin{figure*}[h!]
	\centering
	\begin{subfigure}{0.23\textwidth}
	\centering
	\includegraphics[width=\linewidth]{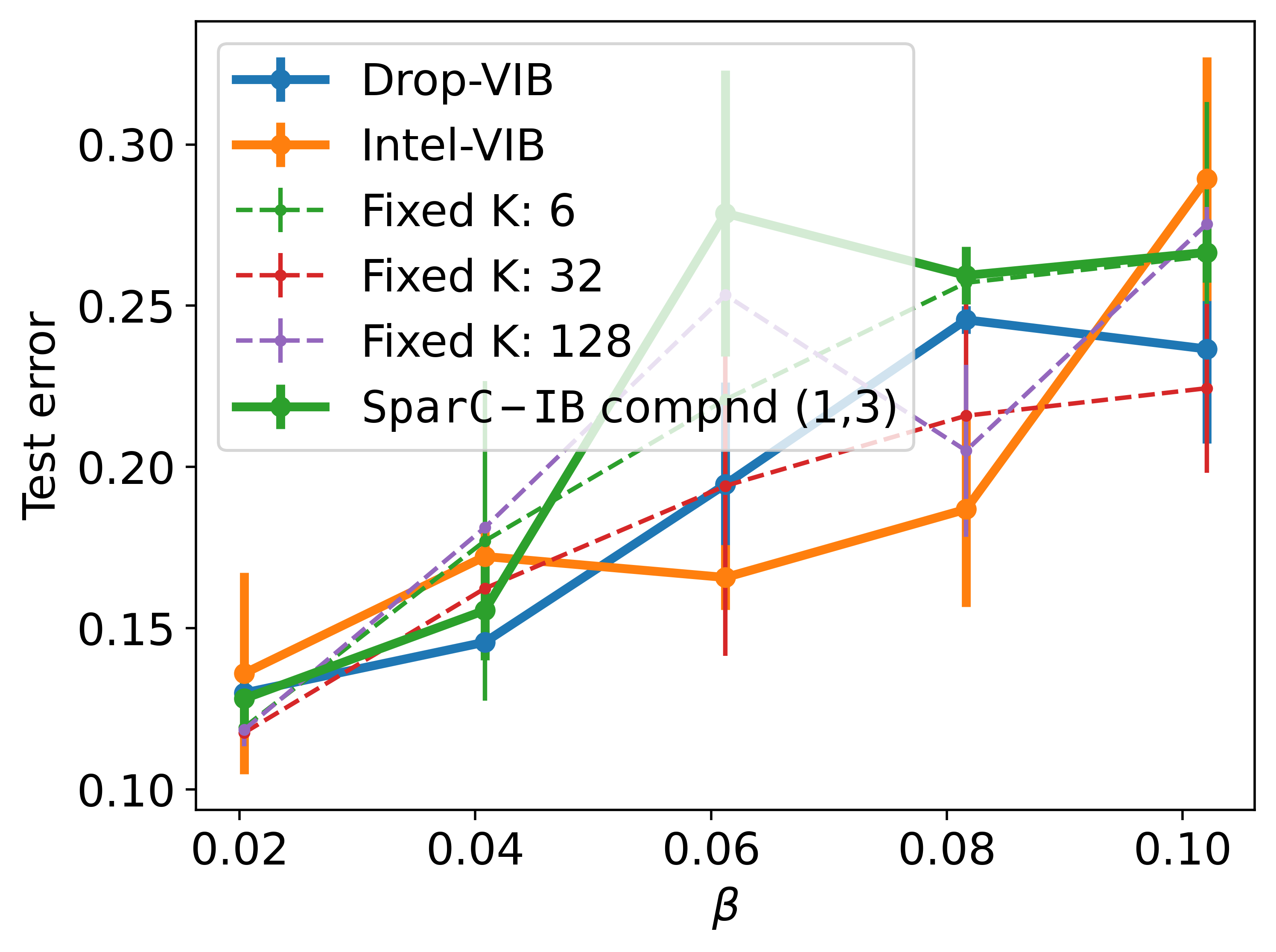}
	\caption{Test error vs. $\beta$ }
	\end{subfigure}
        \hspace{1cm}
	\begin{subfigure}{0.23\textwidth}
	\centering
	\includegraphics[width=\linewidth]{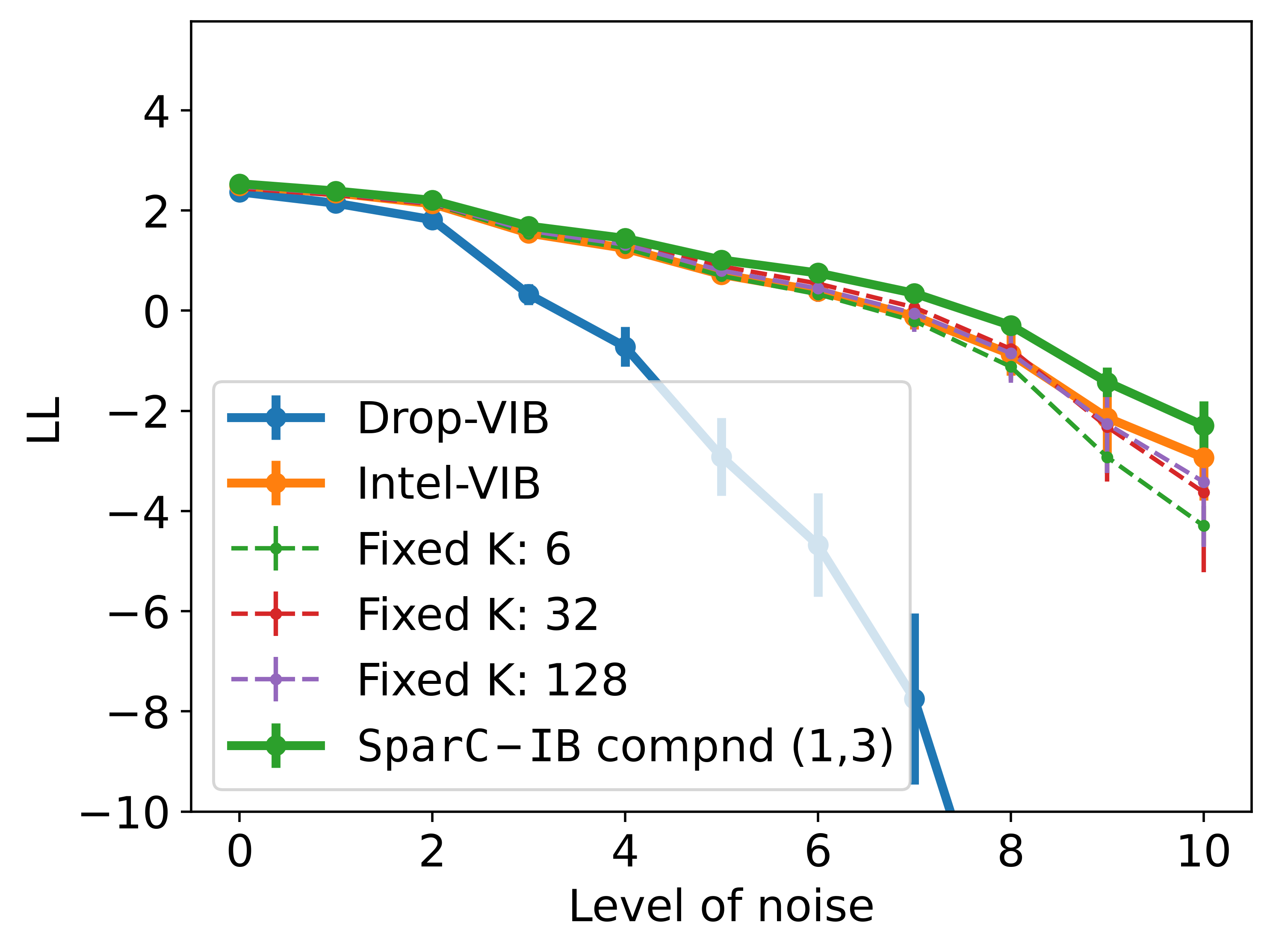}
	\caption{Log likelihood vs. noise}
	\end{subfigure}
         \hspace{1cm}
	\begin{subfigure}{0.23\textwidth}
	\centering
	\includegraphics[width=\linewidth]{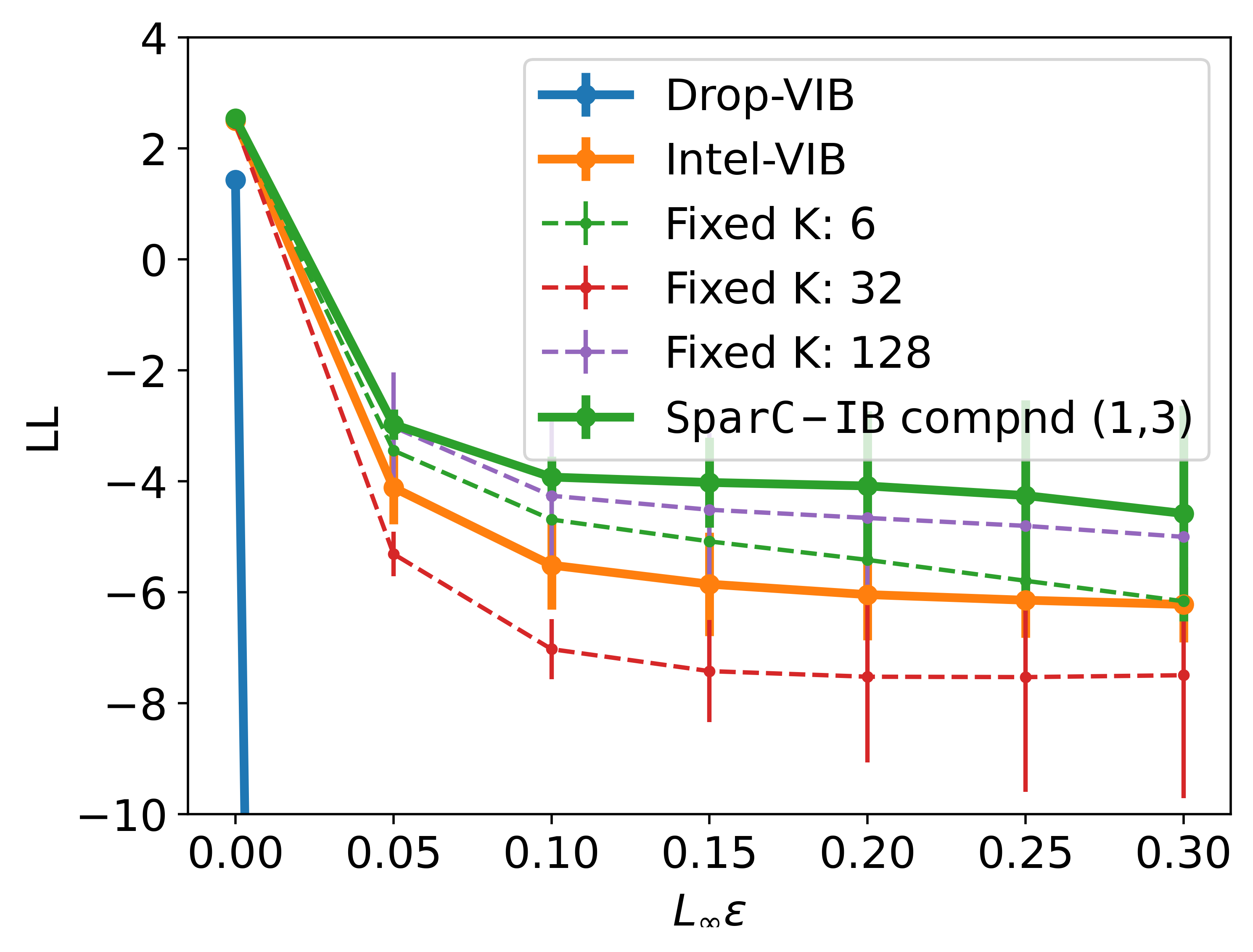}
	\caption{Log likelihood vs. attack}
	\end{subfigure}
        \vspace{-0.5\baselineskip}
	\caption{In- and out-of- distribution performance on CIFAR-10.
         \vspace{-0.15in}}
	\label{fig:Perf_CIFAR}
\end{figure*}

In this section, we present experimental results that compare the performance of \texttt{SparC-IB} with the most recent sparsity-inducing strategies proposed in the literature: the Drop-B model (\cite{kim2021drop}) and InteL-VAE (\cite{miao2022incorporating}). We could not find an opensource implementation for either of the two approaches, and therefore we have coded our own implementation where adapt them in the context of information bottleneck (link to the code base is provided in the Appendix \hyperref[sec::software]{Sec. 6.5}). In this section, these two approaches are called Drop-VIB and Intel-VIB.
In addition, we compare our model with the baseline mean-field Gaussian VIB approach, where the latent dimension is fixed to a single value across all data. Note that, we apply the square transformation (\cite{Kolchinsky19}) to the estimator of $\operatorname{MI}(X;Z)$ for all the methods. 


We evaluated the \texttt{SparC-IB}  model for \emph{in-distribution} prediction accuracy in a supervised classification scenario using the MNIST and CIFAR-10 data, which have a small number of classes, and also the ImageNet data, where the number of classes is large. Furthermore, we evaluated the robustness of \texttt{SparC-IB} trained on these three datasets in \emph{out-of-distribution} scenarios, specifically with rotation, white-box attacks with noise corruptions, and black-box adversarial attacks.

The selection of the Lagrange multiplier $\beta$ controls the amount of information learned from the input (that is, $\operatorname{MI}(X;Z)$) by the latent space. We have chosen a common $\beta$ where ($\operatorname{MI}(X;Z), \operatorname{MI}(Z;Y)$) is close to the \emph{minimum necessary information} or MNI (suggested in \cite{Fischer20}) for all models. MNI is a point in the information plane where $\operatorname{MI}(X;Z) = \operatorname{MI}(Z;Y) = \operatorname{H}(Y)$, where the entropy is indicated by $\operatorname{H}(Y)$. We evaluated the robustness of each model using a single value of $\beta$. The value of $\beta$ we chose to compare the models for MNIST is $\sim 0.08$, for CIFAR-10 is $\sim 0.04$, and for ImageNet is $\sim 0.02$. The choice of $\beta$ is discussed in more detail in the Appendix \hyperref[Informationcurve]{Sec. 6.6}. 

We use the encoder-decoder architecture from \cite{Galvez2020} for MNIST, from \cite{yu2021deep} for CIFAR-10, and from \cite{alemi2016deep} for ImageNet. Note that we learn the parameters of the dimension distribution in both compound and categorical strategies by splitting the encoder network head into two parts, as depicted in \hyperref[BBP_Schematic]{Fig. 2.} Full details of the architectures have been discussed in the Appendix \hyperref[sec::model_arch]{Sec. 6.6}. Furthermore, following \cite{fischer2020ceb}, we pass the mean of the encoder $Z$ to the decoder at the test time to make a prediction.

Prior probabilities act as regularizers in learning the dimension probabilities in both categorical and compound strategies (the third term of $\mathcal{L}_{\texttt{SparC-IB}}(Z,d)$). They also model the prior knowledge or inductive bias that one may have. The prior probability distribution in this case \hyperref[compound]{(8)} can be set by the choice of hyperparameters $(a,b)$.
we evaluate two different cases, $(a,b) = (1,3)$ and (2,2) for both the categorical and compound distribution strategies. The choice $(a,b) = (1,3)$ puts more probability mass on the lower dimensions, and it gradually decays with dimension, whereas $(a,b) = (2,2)$ penalizes models of too high or too low dimensions. The appendix \hyperref[sec::prior_param]{Sec. 6.7} shows the prior probabilities of the dimensions for both choices.


\subsection{Inference on \emph{in-distribution} data}

In this section, we compare the performance of the \texttt{SparC-IB} with Intel-VIB, Drop-VIB, and the vanilla fixed-dimensional VIB approach on the MNIST, CIFAR-10 test set, and ImageNet validation set. We train each model for 5 values of the Lagrange multiplier $\beta$ in the set $(0.02,0.04,0.06,0.08,0.1)$. We calculate the validation set error for each model for these $\beta$ values. Since increasing $\beta$ penalizes the amount of information retained by the latent space about the inputs, we expect the error to increase as $\beta$ increases.      


For MNIST, we find that, across $\beta$, the compound distribution prior with $(a,b) = (2,2)$ performs best in terms of in-distribution prediction accuracy across $\beta$ as compared to other \texttt{SparC-IB} choices, fixed-dimensional VIB approaches and Drop-VIB and Intel-VIB, as shown in \hyperref[fig:Perf_MNIST]{Fig. 3(a)} . For CIFAR-10 data, we observe that compound strategy with prior $(a,b)=(1,3)$ has the best accuracy compared to other \texttt{SparC-IB} choices at MNI, fixed-dimensional VIB models and Intel-VIB but is slightly lower than Drop-VIB (numbers are shown in \hyperref[tab:test_MNISTCIFAR]{Table 2}). However, we find from \hyperref[tab:test_MNISTCIFAR]{Table 2} that \texttt{SparC-IB} compound (1,3) has the highest log-likelihood at MNI among the other approaches.
\begin{table}[h!]
 \centering
 \resizebox{0.9\columnwidth}{!}{   \begin{tabular}{|c|c|c|c|c|}
    \hline
    \multirow{2}{1cm}{\textbf{Methods}} & \multicolumn{2}{c|}{\textbf{MNIST}} & \multicolumn{2}{c|}{\textbf{CIFAR-10}} \\
    \cline{2-5}
    & \textbf{Acc} \% & \textbf{LL} & \textbf{Acc} \% & \textbf{LL} \\
	\hline
	\texttt{SparC-IB} & \textbf{98.65} (0.001)  & \textbf{3.24} (0.004)  & 84.44 (0.015) & \textbf{2.53} (0.010)\\
	Drop-VIB & 98.25 (0.000) & 3.12 (0.003) & \textbf{85.43} (0.004) & 2.37 (0.015)\\
	Intel-VIB & 98.51 (0.001) & 3.23 (0.007) & 82.77 (0.010) & 2.49 (0.063)\\
	Fixed K: 6 & 98.27 (0.001) & 3.22 (0.004) & 82.29 (0.050) & 2.49 (0.107)\\
	Fixed K: 32 & 98.54 (0.001) & 3.23 (0.003) & 83.76 (0.012) & 2.44 (0.013)\\
	Fixed K: 128 & 85.28 (0.165) & 2.85 (0.358) & 81.88 (0.001) & 2.48 (0.036)\\
    \hline
    \end{tabular}
    }
    \caption{In-distribution performance of all methods in terms of accuracy and log-likelihood (SD in the parenthesis) at MNI for MNIST and CIFAR-10 (maximum for each column highlighted). \texttt{SparC-IB} performs as good as the best performing model in terms of both metrics.}
    \label{tab:test_MNISTCIFAR}
\end{table}


For Imagenet data, we observe that the compound strategy with 
prior $(a,b)=(2,2)$ has the best validation accuracy compared 
to other \texttt{SparC-IB} choices. Furthermore, the in-distribution
performance is at least as good as the fixed-dimensional VIB model (\hyperref[tab:test_ImageNet]{Table 3}), but it 
has a slightly lower accuracy compared to the Drop-VIB and Intel-VIB.
The test error for Intel-VIB is very high when $\beta > 0.04$. This behavior is due to the fact that the dimension selector in Intel-VIB (Section 6.3 in \cite{miao2022incorporating}) is pruning almost all values of the latent allocation vector $A$ for values of $\beta > 0.04$.
\begin{table}[h!]
 \centering
 \resizebox{0.5\columnwidth}{!}{   
 \begin{tabular}{|c|c|c|}
    \hline
    \multirow{2}{1cm}{\textbf{Methods}} & \multicolumn{2}{c|}{\textbf{ImageNet}} \\
    \cline{2-3}
    & \textbf{Acc} \% & \textbf{LL} \\
	\hline
	\texttt{SparC-IB} & 79.71 (0.001)  & 8.48 (0.007) \\
	Drop-VIB & 79.86 (0.000) & 8.19 (0.002)\\
	Intel-VIB & \textbf{79.88 (0.000)} & \textbf{8.53 (0.003)} \\
	Fixed K: 1024 & 79.88 (0.000) & 8.52 (0.005) \\
    \hline
    \end{tabular}
    }
    \caption{In-distribution performance of all methods in terms of accuracy and log-likelihood (SD in the parenthesis) at MNI for ImageNet (maximum for each column highlighted). All models are close in terms of both metrics (especially log-likelihood).}
    \label{tab:test_ImageNet}
\end{table}


\subsection{Inference on \emph{out-of-distribution} data}

We consider three out-of-distribution scenarios to measure the robustness of our approach and compare it with vanilla VIB with fixed latent dimension capacity, as well as with other sparsity-inducing strategies. The first is a white-box attack, in which we systematically introduce shot noise corruptions into the test data \cite{mu2019mnistc,CIFAR-C}. The second is a rotation transform (only for MNIST data).
The third is the black-box adversarial attack simulated using the projected gradient descent (PGD) strategy \cite{madry2019deep}. We use the log-likelihood metric to compare the out-of-distribution performance of the methods (\cite{ovadia2019can}). Comparison in terms of other metrics are included in the Appendix \hyperref[sec:OOD_metrics]{Sec. 6.8}.


\subsubsection{White-box attack/Noise corruption}
White-box attack or noise corruption is generated by adding shot noise. Following \cite{mu2019mnistc}, Poisson noise is generated pixel-wise and added to the test images for both MNIST and CIFAR-10. The levels of noise along the horizontal axis of panel (b) of \hyperref[fig:Perf_MNIST]{Fig. 3} and \hyperref[fig:Perf_CIFAR]{Fig. 4} represent an increasing degree of noise added to the images of the validation set. \hyperref[Noise_level]{Fig. 5} shows the impact of adding increasing levels of noise to a sample validation set MNIST image. For our approach and each of the five models that are being compared, we plot the log-likelihood as a function of the level of noise to assess the robustness of these approaches. We find that with both MNIST (panels (b) in \hyperref[fig:Perf_MNIST]{Fig. 3}) and CIFAR-10 (panels (b) in \hyperref[fig:Perf_CIFAR]{Fig. 4}), for each of these three metrics the \texttt{SparC-IB} outperforms all the other compared approaches. For ImageNet (panel (b) in \hyperref[fig:BB_ImageNet]{Fig. 5}), we find that the Drop-VIB has higher likelihood than our approach possibly due to a large amount of input information (estimated $\operatorname{MI}(X;Z) = 57.56$ for Drop-VIB and = 10.35 for \texttt{SparC-IB}) learned in the latent space.

\subsubsection{Rotation}
In this scenario, we evaluate the models trained on MNIST data with increasingly rotated digits following~\cite{ovadia2019can}, 
which simulates the scenario of data that are moderately out of distribution to the data used for training the model. We show the results of the experiments in panel (c) in \hyperref[fig:Perf_MNIST]{Fig. 3}. We can see that \texttt{SparC-IB} with the compound distribution prior outperforms the rest of the five models in terms of log-likelihood, we also note that the performance of the Drop-VIB has the lowest in this scenario.

\subsubsection{Adversarial Robustness}
\begin{figure}
	\centering
	\begin{subfigure}{0.15\textwidth}
	\centering
	\includegraphics[width=\linewidth]{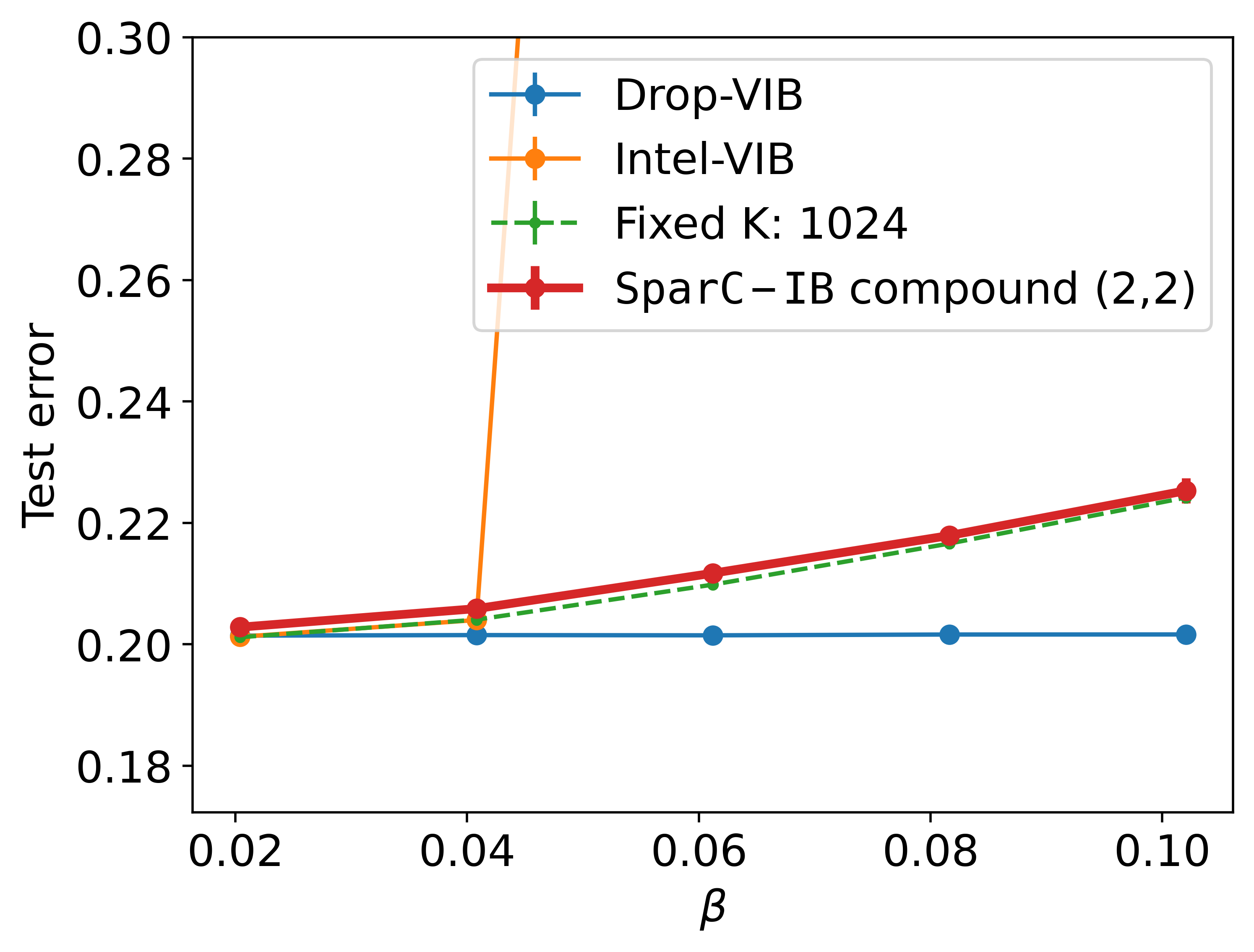}
	\caption{Test error vs. $\beta$ value}
	\end{subfigure}
        \hfill
	\begin{subfigure}{0.15\textwidth}
	\centering
	\includegraphics[width=\linewidth]{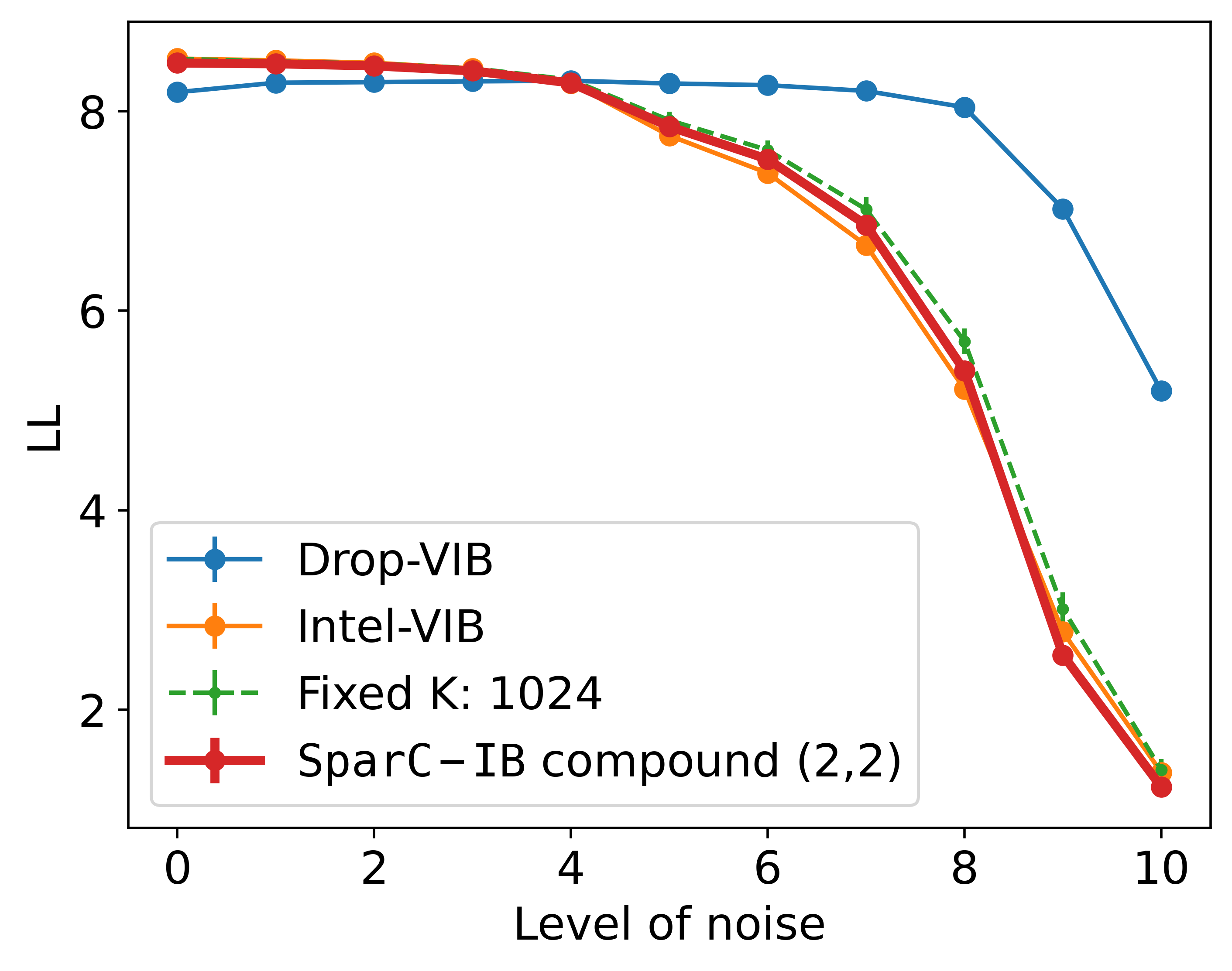}
	\caption{Log likelihood vs. noise}
	\end{subfigure}
        \hfill
	\begin{subfigure}{0.15\textwidth}
	\centering
	\includegraphics[width=\linewidth]{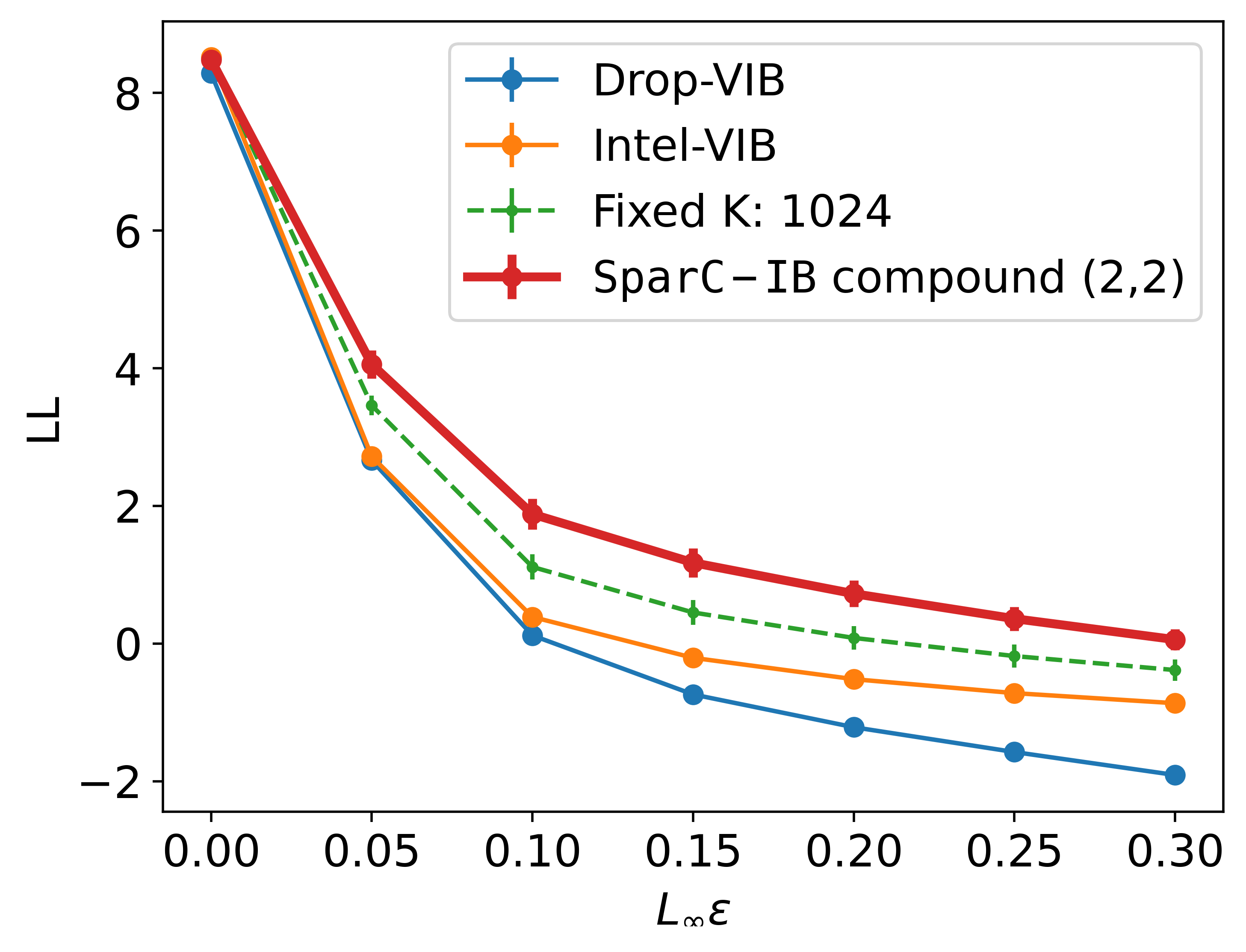}
	\caption{Log likelihood vs. attack radius}
	\end{subfigure}
	\caption{In- and out-of- distribution results on ImageNet.}
	\label{fig:BB_ImageNet}
\end{figure}
Multiple approaches have been proposed in the literature to assess the adversarial robustness 
of a model~\cite{tramer2020adaptive}. Among them, perhaps the most commonly adopted approach is to test the accuracy in adversarial examples
that are generated by adversarially perturbing the test data. This perturbation is in the same spirit as the noise
corruption presented in the preceding section but is designed to be more catastrophic by adopting a black-box
attack approach that chooses a gradient direction of the image pixels at some loss and then takes a single step 
in that direction. The projected gradient descent is an example of such an adversarial attack. Following~\cite{alemi2016deep},
we  evaluate the model robustness for the PGD attack with 10 iterations. We use the $L_{\infty}$ norm to measure the size of the perturbation, which in this case is a measure of the largest single change in any pixel. Log-likelihood as a function of the perturbed $L_{\infty}$ distance for MNIST (panel (d) in \hyperref[fig:Perf_MNIST]{Fig. 3}), for CIFAR-10 (panel (c) in \hyperref[fig:Perf_CIFAR]{Fig. 4}), and for ImageNet (panel (c) in \hyperref[fig:BB_ImageNet]{Fig. 5}) \emph{show that the} \texttt{SparC-IB} \emph{approach provides the highest log-likelihood across the attack radius in all three datasets}.

\subsection{Analysis of the latent space}
A key property of \texttt{SparC-IB} is the ability to jointly learn the latent allocation and the dimension of the latent space for each data point. In this section, our aim is to disentangle the information learned in the latent space of the \texttt{SparC-IB} approach by analyzing the posterior distribution of the dimension variable and the information learned in the latent allocation vector for MNIST data. Similar analyses for CIFAR-10 and ImageNet are in the Appendix \hyperref[sec::latentspace]{Sec. 6.9}.

\subsubsection{Flexible dimension of the latent space}

A distinct feature of the proposed approach is the flexibility enabled by the sparsity-inducing prior for learning a data-dependent dimension distribution. To demonstrate this, for a compound model with $(a,b)=(2,2)$ in \hyperref[fig:Dimension_vs_digit_MNIST]{Fig. 1} we show the distribution of posterior modes of dimension distribution across data points, aggregated per MNIST digit. We see that, in fact, each digit, on average, preferred to have a different latent dimension. We further note that the mode values depicted by the plot for digits 5 and 8 are farther away from the rest of the digits. Note that the pattern in \hyperref[fig:Dimension_vs_digit_MNIST]{Fig. 1} is for a single seed. Although the pattern in dimension distribution modes changes across the seeds, the separation between the classes remains (see \hyperref[sec::latentspace]{Sec. 6.9.1} in the Appendix for details). We also observe a similar separation of the latent dimensionality across classes with CIFAR-10 and ImageNet data. In our experiments, we have observed that the robustness performance improves for seeds with higher separation of the modes between classes.

\subsubsection{Analyzing information content}
\label{sec::InfoCont}
\begin{figure}
    \centering
    \includegraphics[width=0.5\linewidth]{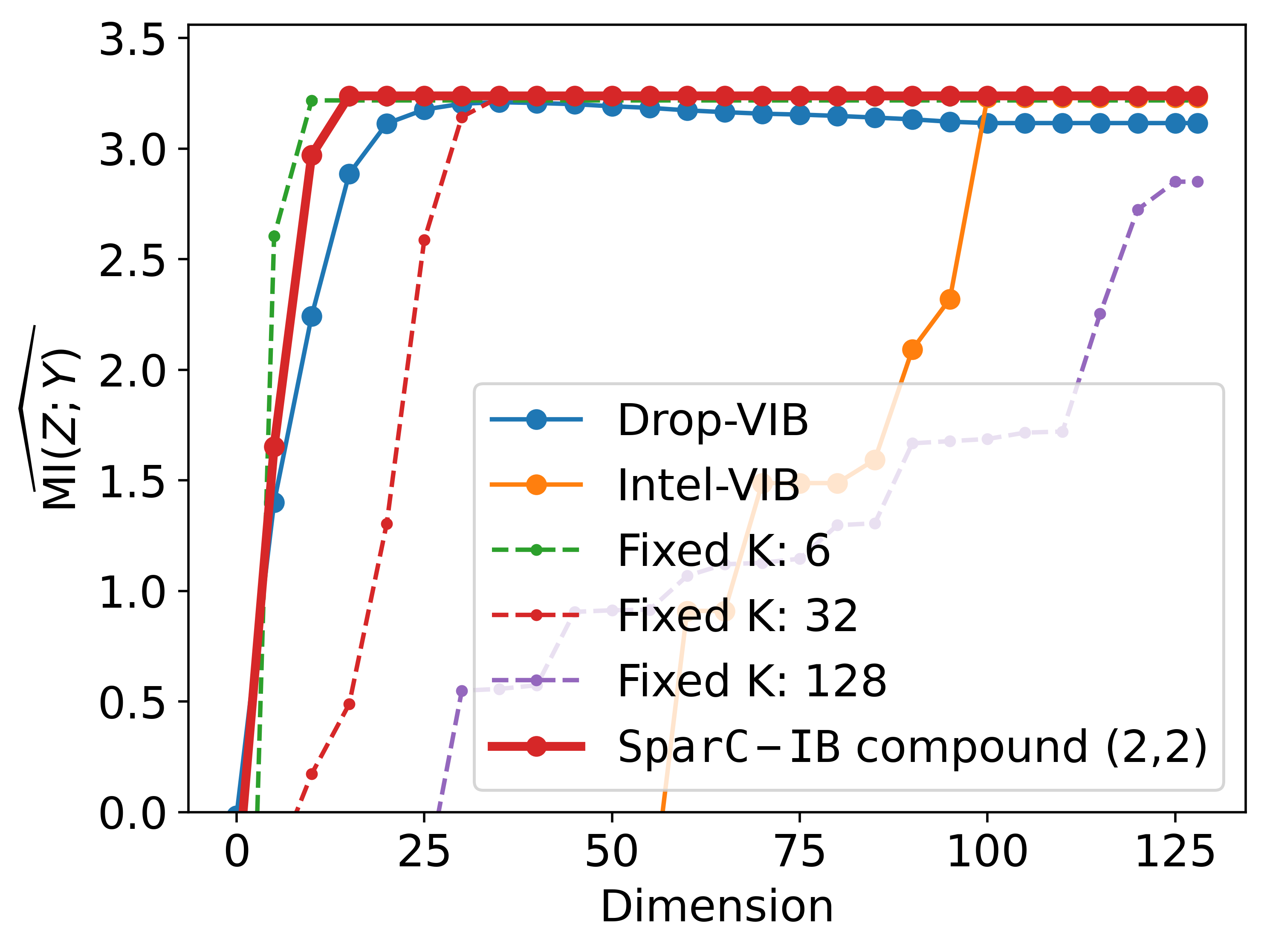}
    \caption{Information content plot for different latent dimensions (averaged across seeds). We observe that \texttt{SparC-IB} learns the maximum information in a small dimensional latent space.}
    \label{fig:InfoCont_MNIST}
\end{figure}
The \texttt{SparC-IB} prior in \hyperref[prior]{(6)} induces an ordered selection of the dimensions of the latent allocation vector $A_n$ based on the dimension $d_n$. In \hyperref[fig:InfoCont_MNIST]{Fig. 6}, we plotted the estimated mutual information or $\widehat{\operatorname{MI}(Z;Y)}$ against the increasing dimension of the latent space (in increment of 5) for all models in MNIST. For a given dimension $d$, $\widehat{\operatorname{MI}(Z;Y)} = \frac{1}{N} \sum_{n=1}^{N} \log q(Y_n|Z_n = \mu_n (X) \circ \gamma(d))$, where the first $d$ coordinates of $\gamma(d)$ are 1s and the rest are 0s. We observe that the \texttt{SparC-IB} model has been able to code the learned information in the first few ($\sim$ 15) dimensions of the latent space. In contrast, we notice that the fixed-dimensional VIB models, Drop-VIB, and the Intel-VIB require close to the full latent space to encode similar information level. This characteristic perhaps hinders these models to achieve good robustness performance consistently on all data. However, we believe that further investigation is needed to establish this claim. For CIFAR-10, we observed the same characteristics of the information content plot in \hyperref[fig:InfoCont_MNIST]{Fig. 6} whereas for ImageNet the information plateaus at a larger dimension than \hyperref[fig:InfoCont_MNIST]{Fig. 6} (see \hyperref[InfoContentCIFARandIMNet]{Fig. A8} in the Appendix).       

\subsubsection{Visualizing the latent dimensions}
\label{sec::VizLatent}
\begin{figure}
    \centering
    \includegraphics[width=\linewidth]{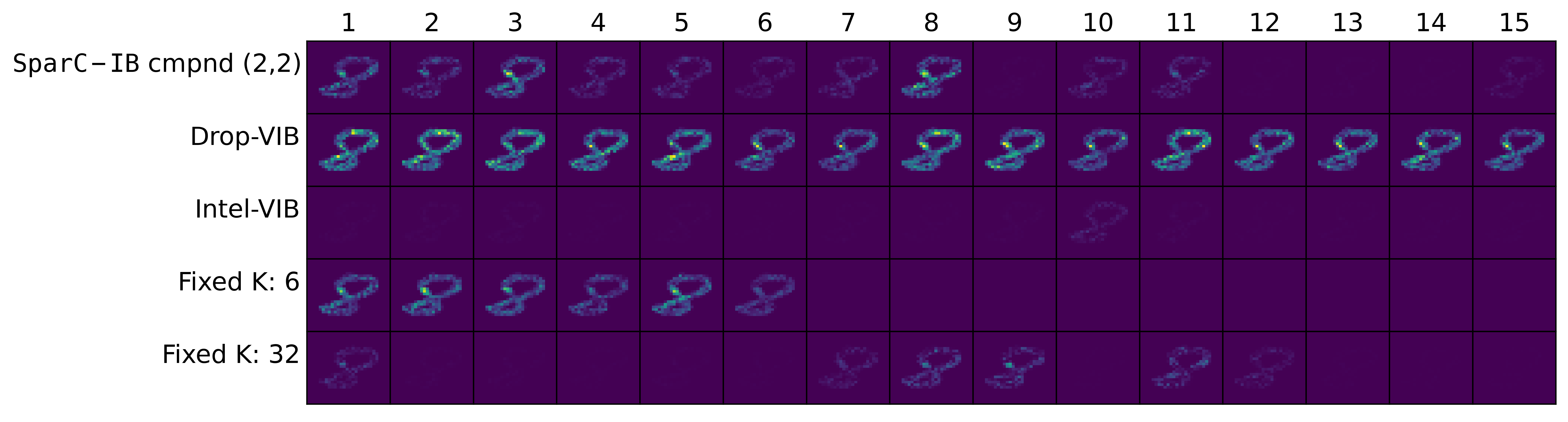}
    \caption{Plot of pixel importance in encoding the latent allocation mean (averaged across seeds) for a sample from the MNIST test data. We observe that \texttt{SparC-IB} learns important features of the latent space in the first few dimension of the latent space.}
    \label{fig:FltoNeuron_MNIST}
\end{figure}
In deep neural networks, the conductance of a hidden unit, proposed in \cite{Dhamdhere2018}, is the flow of information from the input through that unit to the predictions. In this spirit, we measure the importance scores of individual pixels in the dimensions of the latent mean allocation vector $\mu$ for the MNIST data. The computation details have been added in the Appendix \hyperref[sec::calc_ImpScore]{Sec. 6.9.4}. In \hyperref[fig:FltoNeuron_MNIST]{Fig. 7}, we plotted these measures for different methods (averaged across seeds) for the first 15 dimensions of $\mu$ for a sample from the MNIST test data. We observe that \texttt{SparC-IB} encodes important features in the first few dimensions of the latent space. For Intel-VIB and VIB with K = 32, we see that the pixel information is spread over a large set of dimensions of the latent space. For Drop-VIB, we notice that it learns a lot of information in all the 15 dimensions. \hyperref[fig:FltoNeuron_MNIST]{Fig. 7} also helps to explain the information jumps in \hyperref[fig:InfoCont_MNIST]{Fig. 6}. We notice that the jump in the information content occurs when we include the dimensions that have learned important pixel information, e.g. for \texttt{SparC-IB} dimensions 3 and 8. Note that the Intel-VIB and fixed-dimensional VIB models with high dimensions have many dimensions with very little information about the input. Although \hyperref[fig:FltoNeuron_MNIST]{Fig. 7} exhibits some features of the digit learned by the latent space (e.g., the middle part of the digit has high importance for \texttt{SparC-IB}), in our experiments, we have not been able to extract meaningful features in the latent dimensions that are common across all the digits. In our view, this demands further investigation.       

\section{Conclusions}
\label{conclusion}


In summary, we propose the \texttt{SparC-IB} method in this paper which models the latent variable and its dimension through a Bayesian spike-and-slab categorical prior and derived a variational lower bound for efficient Bayesian inference. This approach accounts for the full uncertainty in the latent space by learning a joint distribution of the latent variable and the sparsity. We compare our approach with commonly used fixed-dimensional priors, as well as using those sparsity-inducing strategies previously proposed in the literature through experiments on MNIST, CIFAR-10, and ImageNet in both the in-distribution and out-of-distribution scenarios (such as noise corruption, adversarial attacks, and rotation). We find that our approach obtains as good accuracy and robustness as the best performing model in all the cases and in some cases it outperforms the other models. This is important because we found that other VIB algorithms considered performed well on a few datasets but have significantly poor performance on the others.  In addition, we show that enabling each data to learn their own dimension distribution leads to separation of dimension distribution between output classes, thus substantiating that latent dimension varies class-wise. Furthermore, we find that the \texttt{SparC-IB} approach provides a compact latent space where it learns important data features in the first few dimensions of the latent space, which is known to lead to superior robustness properties. 



There are several avenues for future research based on the proposed model \texttt{SparC-IB}. Since the latent dimensionality of the data is modeled through a Bayesian spike-and-slab prior with a categorical spike distribution over the dimension, an interesting avenue for future research could be to find a rich class of hierarchical priors or a non-parametric stick-breaking prior. Another direction might be to explore other approaches to data-driven mutual information estimation to tighten the lower bound of $\operatorname{MI}(X;Z)$.

\bibliography{sample_paper}
\bibliographystyle{plain}




\newpage

\section{Appendix}
\beginsupplement
\subsection{Model architectures and Hyperparameter settings}
\label{sec::model_arch}
For the proposed \texttt{SparC-IB} model, we need to select the neural architecture to be used for the latent space mean and variance, as well as the dimension encoder that learns categorical probabilities or compound distribution parameters. To learn the parameters of the dimension distribution in both compound and categorical strategies, we split the head of the encoder network into two parts, as depicted in \hyperref[BBP_Schematic]{Fig. 2}. The first part estimates the parameters of the latent allocation variable $A$ and the second part estimates the parameters of the dimension variable $d$. For MNIST, we follow \cite{Galvez2020} and use an MLP encoder with three fully connected layers, the first two of dimensions $800$ and ReLu activation, and the last layer with dimension $2K+2$ for the compound strategy and $3K$ for the categorical strategy. The decoder consists of two fully connected layers, the first of which has dimension $800$ and the second predicts the softmax class probabilities in the last layer. For CIFAR-10, following \cite{yu2021deep}, we adopt a VGG16 encoder and a single-layered neural network as decoder. We choose the final sequential layer of VGG16 as the bottleneck layer that outputs the parameters of the latent space. 

For ImageNet, we crop the images at its center to make them 299 $\times$ 299 pixels and normalize them to have mean = (0.5, 0.5, 0.5) and standard deviation = (0.5, 0.5, 0.5). We have followed the implementation of \cite{alemi2016deep} where we transform the ImageNet data with a pre-trained Inception Resnet V2 (\cite{szegedy16}) network without the output layer. Under this transformation, the original ImageNet images reduce to a 1534-dimensional representation, which we used for all our results. Following \cite{alemi2016deep}, we use an encoder with two fully connected layers, each with 1024 hidden units, and a single-layer decoder architecture.

For comparison with other sparsity-inducing approaches, we chose the two most recent works: the Drop-B model (\cite{kim2021drop}) and the InteL-VAE (\cite{miao2022incorporating}). The Drop-B implementation requires a feature extractor. For all three data sets, we choose the same architecture for the feature extractor as the encoder of \texttt{SparC-IB} until the final layer, which has $K$ dimensions. We assume the same decoder architecture for Drop-B as for \texttt{SparC-IB}. Additionally, for Drop-B, the $K$ Bernoulli probabilities are trained with the other parameters of the model. For InteL-VAE, the encoder and decoder architectures are chosen to be the same as in \texttt{SparC-IB} for all data sets. In addition, this model requires a dimension selector (DS) network. Following the experiments in \cite{miao2022incorporating}, we select three fully connected layers for the DS network with ReLu activations between, where the first two layers have dimension 10 and the last layer has dimension $K$. We fix $K$ (that is, the prior assumption of dimensionality) to be 100 for MNIST and CIFAR-10 and 1024 for the ImageNet data.

The workflow of \texttt{SparC-IB} overlaps with the standard VIB when encoding the mean and sigma of the full dimension of the latent variable. Furthermore, \texttt{SparC-IB} encodes categorical probabilities and then draws samples from a categorical distribution. Unlike the reparameterization trick (\cite{kingma2014autoencoding}) for Gaussian variables, there does not exist a differentiable transformation from categorical probabilities to the samples. Therefore, we use the Gumbel-Softmax approximation (\cite{maddison2017concrete}, \cite{jang2017categorical}) to draw categorical samples. We apply the transformation in \hyperref[prior]{Eq. 4} to the samples and take the element-wise product with the Gaussian samples before passing it to the decoder. Note that there exists other differentiable reparameterization of the discrete samples, e.g., the Gapped Straight-Through (GST) estimator \cite{Fan2022}. However, in our experiments, the use of the Gumbel-Softmax approximation has led to a lower loss value than that of the GST.

\begin{figure}
    \centering
    \includegraphics[width = 0.4\linewidth]{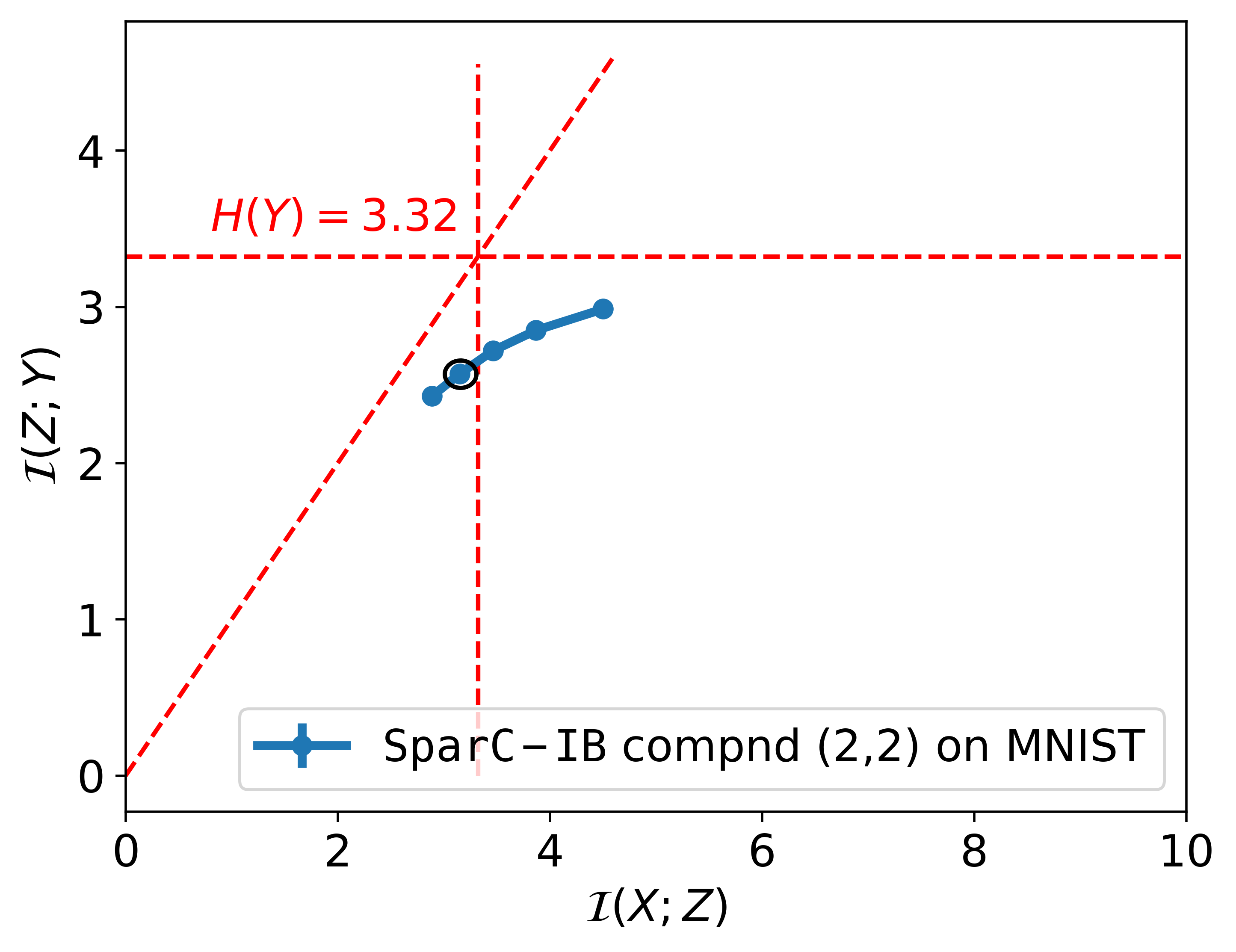}
    \includegraphics[width = 0.4\linewidth]{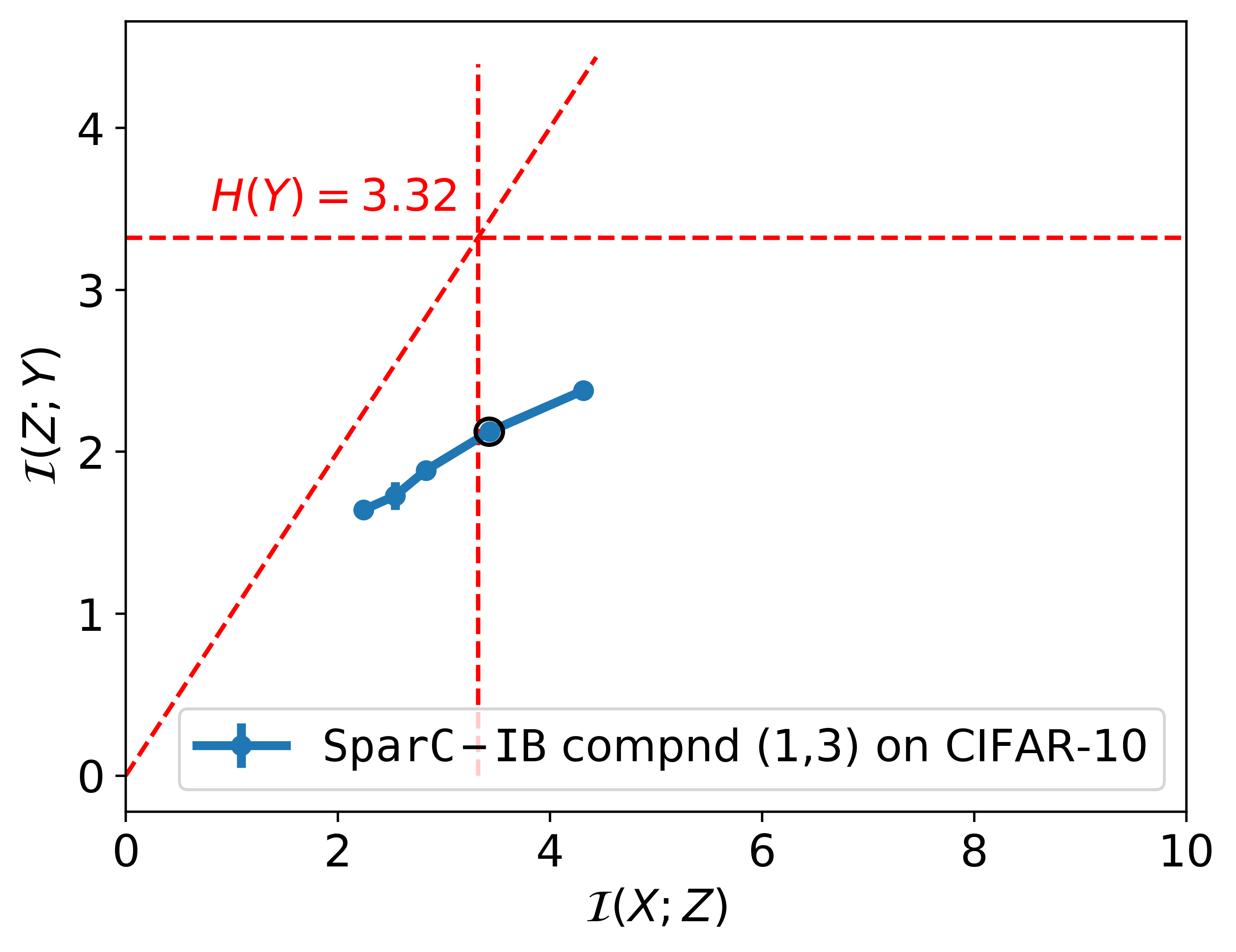}
    \includegraphics[width = 0.4\linewidth]{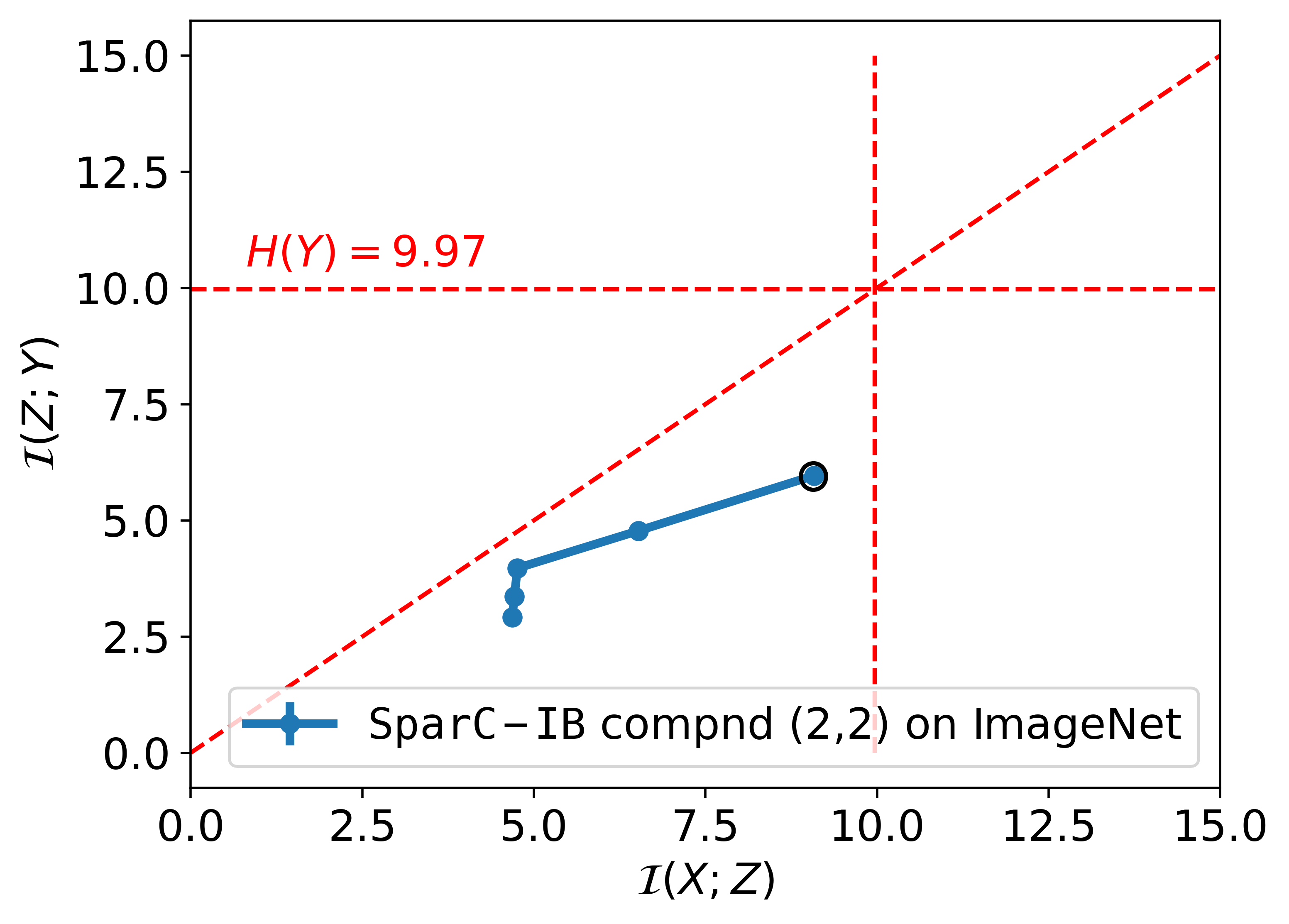}
    \caption{Information curve on MNIST, CIFAR-10, and ImageNet.}
    \label{Informationcurve}
\end{figure}

Fitting deep learning models involves several key hyperparameters. In \hyperref[hyperparams]{Table A1}, we provide the necessary hyperparameters for training and evaluation of all fitted models in three data sets.
\vspace{0.5cm}
\begin{table}[h!]
\center
\resizebox{0.9\columnwidth}{!}{
\begin{tabular}{|c|c|c|c|}
\hline
Hyperparameters & MNIST & CIFAR-10 & ImageNet\\
\hline
Train set size & 60,000 & 50,000 & 128,1167\\

Validation set size & 10,000 & 10,000 & 50,000\\

\# epochs & 100 & 400 & 200\\

Training batch size & 128 & 100 & 2000 \\

Optimizer & Adam & SGD & Adam \\

Initial learning rate & 1e-4 & 0.1 & 1e-4\\

Learning rate drop & 0.6 & 0.1 & 0.97\\

Learning rate drop steps & 10 epochs & 100 epochs & 2 epochs\\

Weight decay & Not used & 5e-4 & Not used\\
\hline
\end{tabular}
}
\vspace{0.2cm}
\caption{Hyperparameter settings used to model the three data sets.}
\label{hyperparams}
\end{table}

\subsection{Data Augmentation}

For the CIFAR-10 data, we augmented the training data using random transformations. We pad each training set image by 4 pixels on all sides and crop at a random location to return an original-sized image. We flip each training set image horizontally with probability 0.5. Furthermore, we normalize each training and validation set image with mean = $(0.4914, 0.4822, 0.4465)$ and standard deviation = $(0.2023, 0.1994, 0.2010)$.

\subsection{Convergence check}

For CIFAR-10, we observed overfitting after 100 epochs for the Drop-VIB model, where the validation loss started to increase. Therefore, we saved the model at epoch 100 for our robustness analysis. For other models in our experiments on MNIST and CIFAR-10, we observed the convergence of train and validation loss, and we have considered models at the final epoch as the final models. For ImageNet, we saved the model with the lowest validation loss as our final model for all the methods.

\subsection{Evaluation metrics}

 We calculated three evaluation metrics for each method in each scenario: test error, log-likelihood, and Brier score. For prediction in all in- and out-of-distribution scenarios, following \cite{fischer2020ceb}, we have used the mean latent space $\mathbb{E}(Z|X)$ as input to the decoder for all methods. Note that for \texttt{SparC-IB} we calculate the marginal expectation by following.
 \begin{flalign}
 \mathbb{E}(Z|X) &= \mathbb{E}_{q(d|X)} \mathbb{E}_{q(Z|d,X)} (Z) \hat{=} \frac{1}{J} \sum_{j=1}^{J} \mathbb{E}_{q(Z|d=d_j,X)} (Z) & \nonumber \\
 & \quad \text{where } d_j \text{ is a sample from } q(d|X) & \nonumber 
 \end{flalign}
In our experiments, we have fixed $J=10$.

\begin{figure}
    \centering
    \includegraphics[width = 0.4\linewidth]{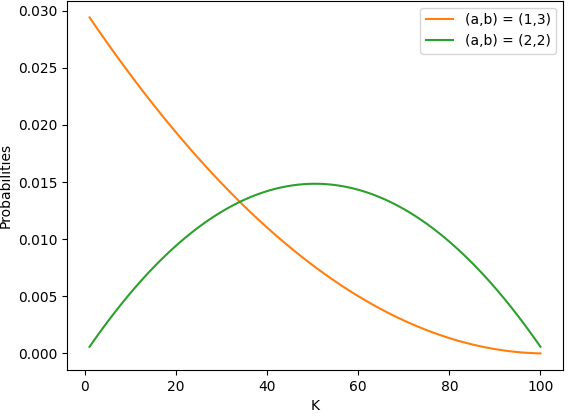}
    \includegraphics[width = 0.45\linewidth]{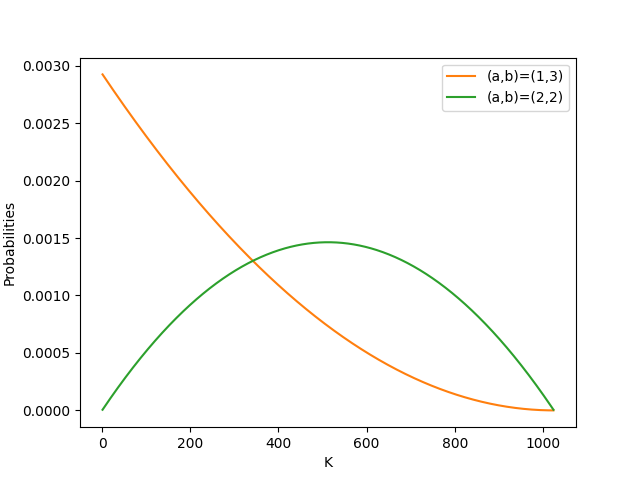}
    \caption{Plot of prior dimension probabilities for choices of $(a,b)$.}
    \label{Prior}
\end{figure}

\subsection{Software and Hardware}
\label{sec::software}
We have forked the code base \url{https://github.com/burklight/convex-IB-Lagrangian-PyTorch.git} that implements the Convex-IB method (\cite{Galvez2020}) using PyTorch. The code to run the models used in the experiments can be found in the following repository \url{https://anonymous.4open.science/r/Sparse-Bayes}. For modeling MNIST, we used NVIDIA V100 GPUs and for CIFAR-10 and ImageNet experiments, we used NVIDIA A100 GPUs. 

\subsection{Information Curve: Selection of $\beta$}
\label{Choice_of_beta}
In the IB Lagrangian, the Lagrange multiplier $\beta$ controls the trade-off between two MI terms. By optimizing the IB objective for different values of $\beta$, we can explore the information curve, which is the plot of $(\operatorname{MI}(X;Z), \operatorname{MI}(Z;Y))$ in the 2-d plane. \hyperref[Informationcurve]{Fig. A1} shows the information curve on the validation set for the models selected for MNIST, CIFAR-10 and ImageNet for the robustness studies in the main article. For the fixed $K$ VIB models, the information curves are similar to \hyperref[Informationcurve]{Fig. A1}. \emph{Minimum necessary information} (\cite{Fischer20}) is a point in the information plane where $\operatorname{MI}(X;Z) = \operatorname{MI}(Z;Y) = \operatorname{H}(Y)$, where the entropy is indicated by $\operatorname{H}(Y)$. For classification tasks, where labels are deterministic given the images, the entropy $\operatorname{H}(Y) = \log_2 n_c$, where $n_c$ is the number of classes. Therefore, for MNIST and CIFAR-10 $\operatorname{H}(Y) = \log_2 10 \sim 3.32$ and for ImageNet $\operatorname{H}(Y) = \log_2 10 \sim 9.97$. Therefore, we choose $\beta \sim 0.08$ for MNIST, $\beta \sim 0.04$ for CIFAR-10, and $\beta \sim 0.02$ for ImageNet, which gives us the closest proximity to MNI. The points are circled in \hyperref[Informationcurve]{Fig. A1}.
\subsection{Selection of prior parameters for \texttt{SparC-IB}}
\label{sec::prior_param}
\hyperref[Prior]{Fig. A2} shows the probabilities of the prior dimension considered for modeling the three data sets.  These prior probabilities are from the compound distribution (\hyperref[compound]{Eq. 6}) with $K=100$ (left figure) and $K=1024$ (right figure).  

\subsection{Performance based on other evaluation metrics on \emph{out-of-distribution} data}
\label{sec:OOD_metrics}

\begin{figure}[h!]
	\centering
	\begin{subfigure}{0.3\linewidth}
	\centering
	\includegraphics[width=\linewidth]{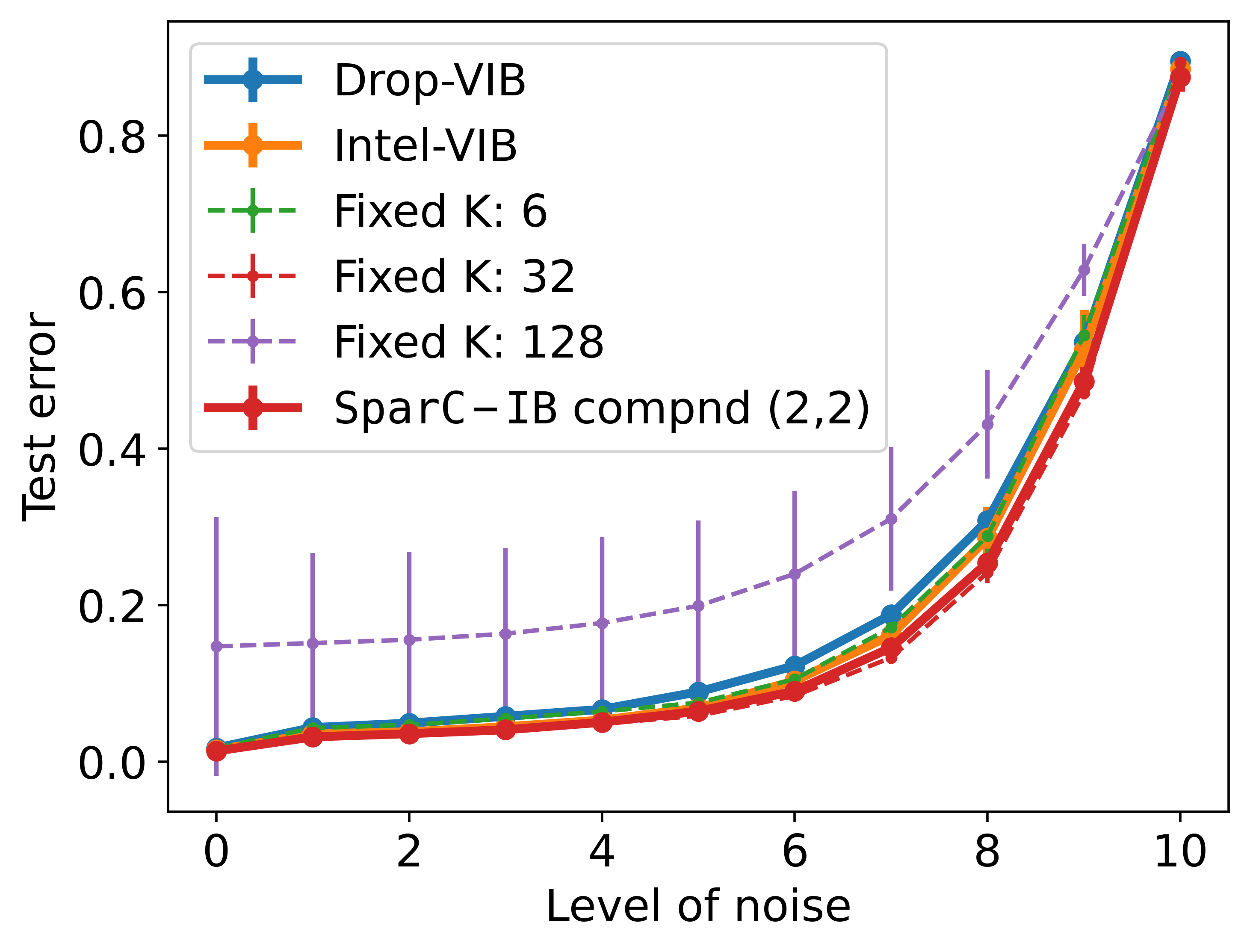}
	\caption{Test error vs noise}
	\end{subfigure}
	\begin{subfigure}{0.3\linewidth}
	\centering
	\includegraphics[width=\linewidth]{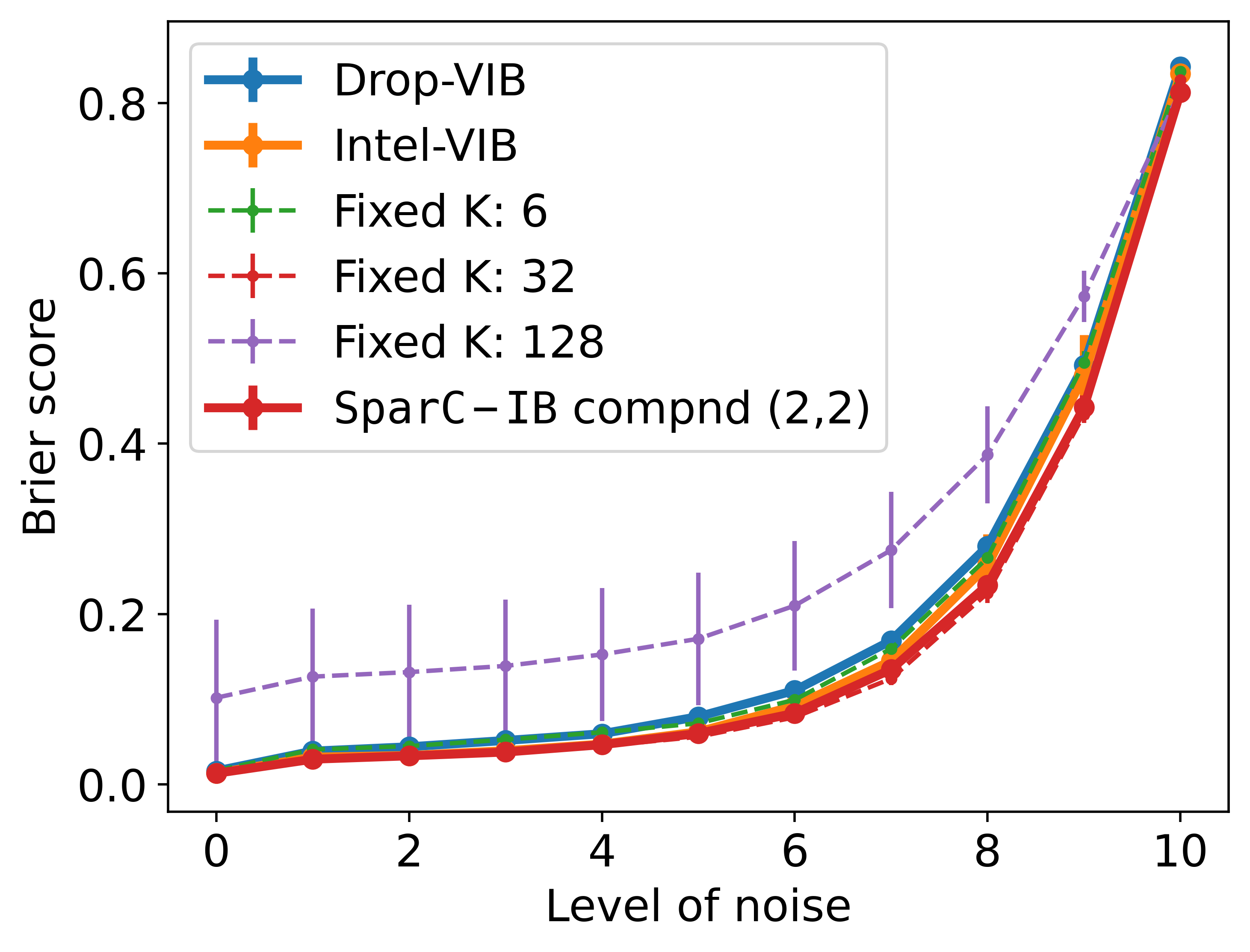}
	\caption{Brier score vs noise}
	\label{brier_v_noise}
	\end{subfigure}
	\begin{subfigure}{0.3\linewidth}
	\centering
	\includegraphics[width=\linewidth]{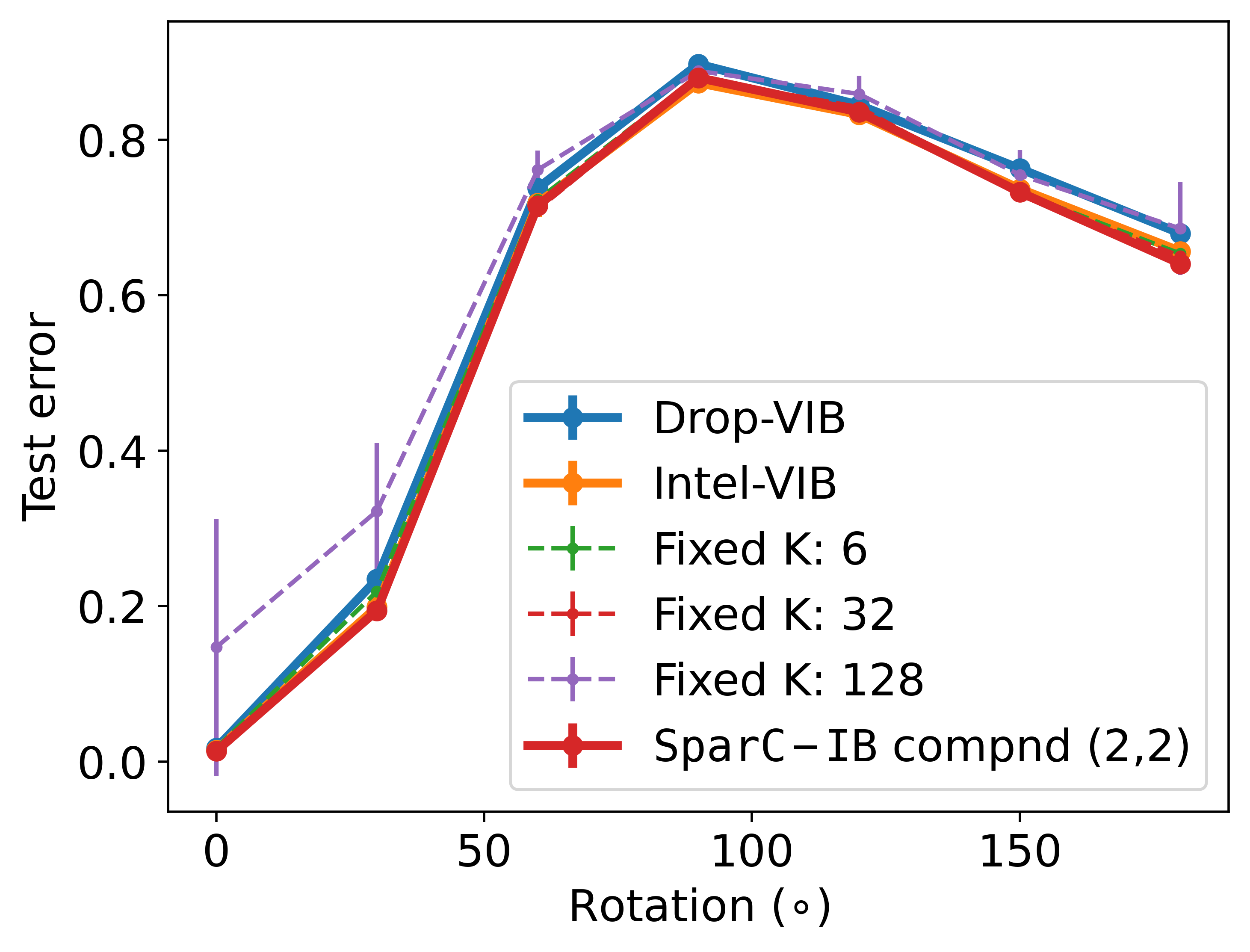}
	\caption{Test error vs rotation}
	\label{Error_v_rotation}
	\end{subfigure}
	\begin{subfigure}{0.3\linewidth}
	\centering
	\includegraphics[width=\linewidth]{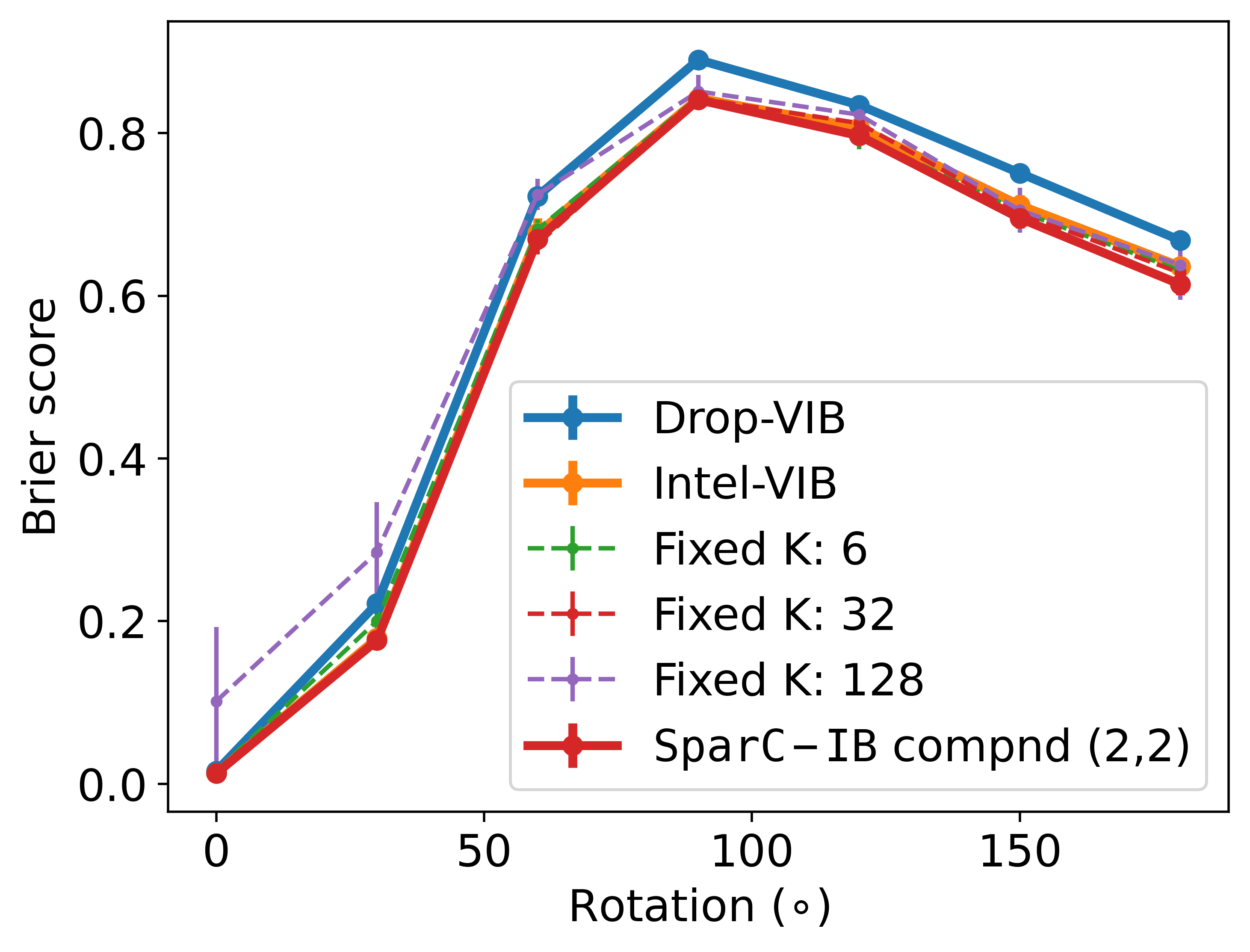}
	\caption{Brier score vs rotation}
	\label{brier_v_rotation}
	\end{subfigure}
	\begin{subfigure}{0.3\linewidth}
	\centering
    \includegraphics[width=\linewidth]{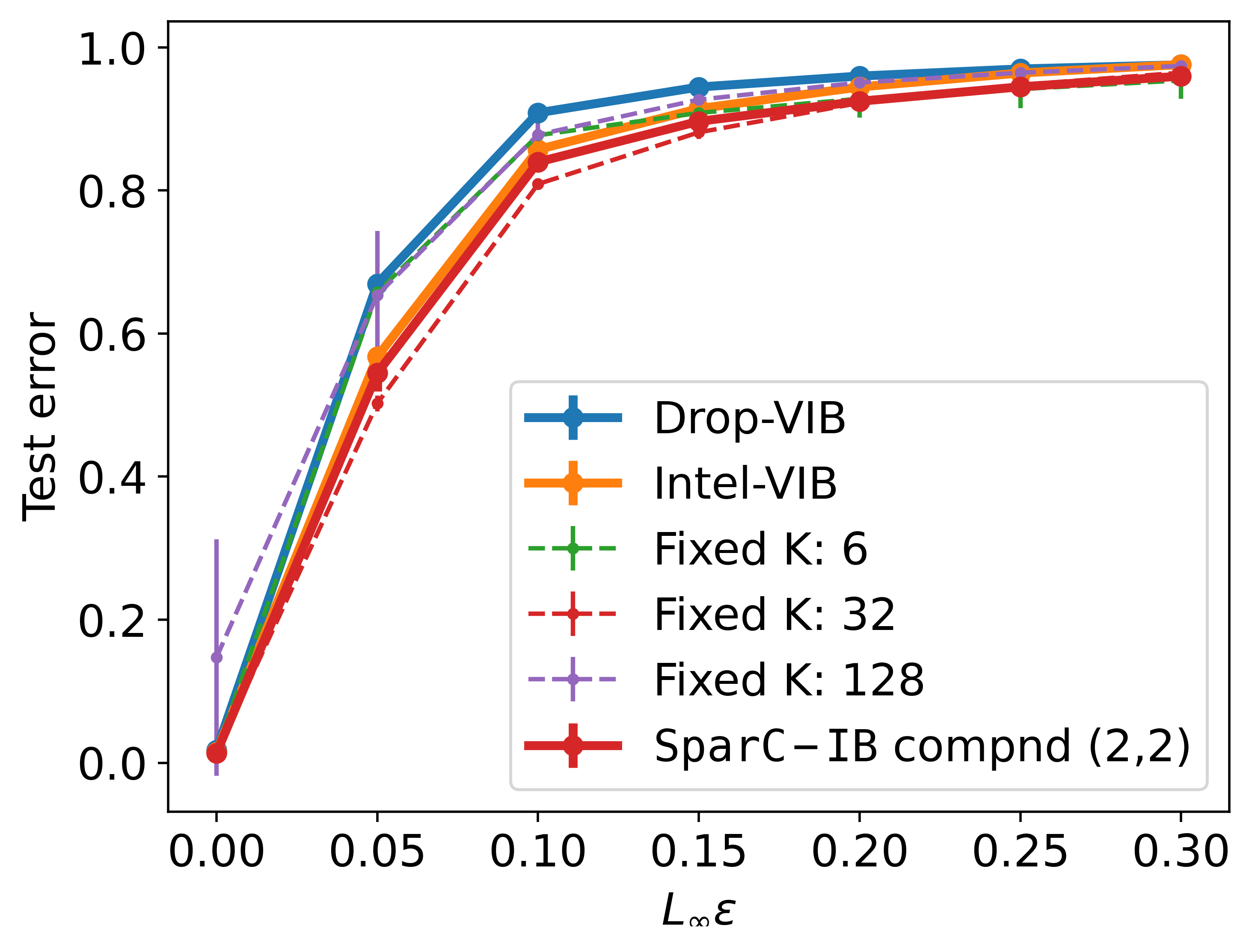}
	\caption{Test error vs $L_\infty$ radius}
	\end{subfigure}
	\caption{Out-of-distribution performance in terms of the test error, and the Brier score on MNIST. We observe that \texttt{SparC-IB} approach with compound strategy and $(a,b)=(2,2)$ (red line) performs as good as the best performing model in all the cases.}
	\label{OOD_noise}
\end{figure}

\begin{figure}
	\centering
	\begin{subfigure}{0.3\linewidth}
	\centering
	\includegraphics[width=\linewidth]{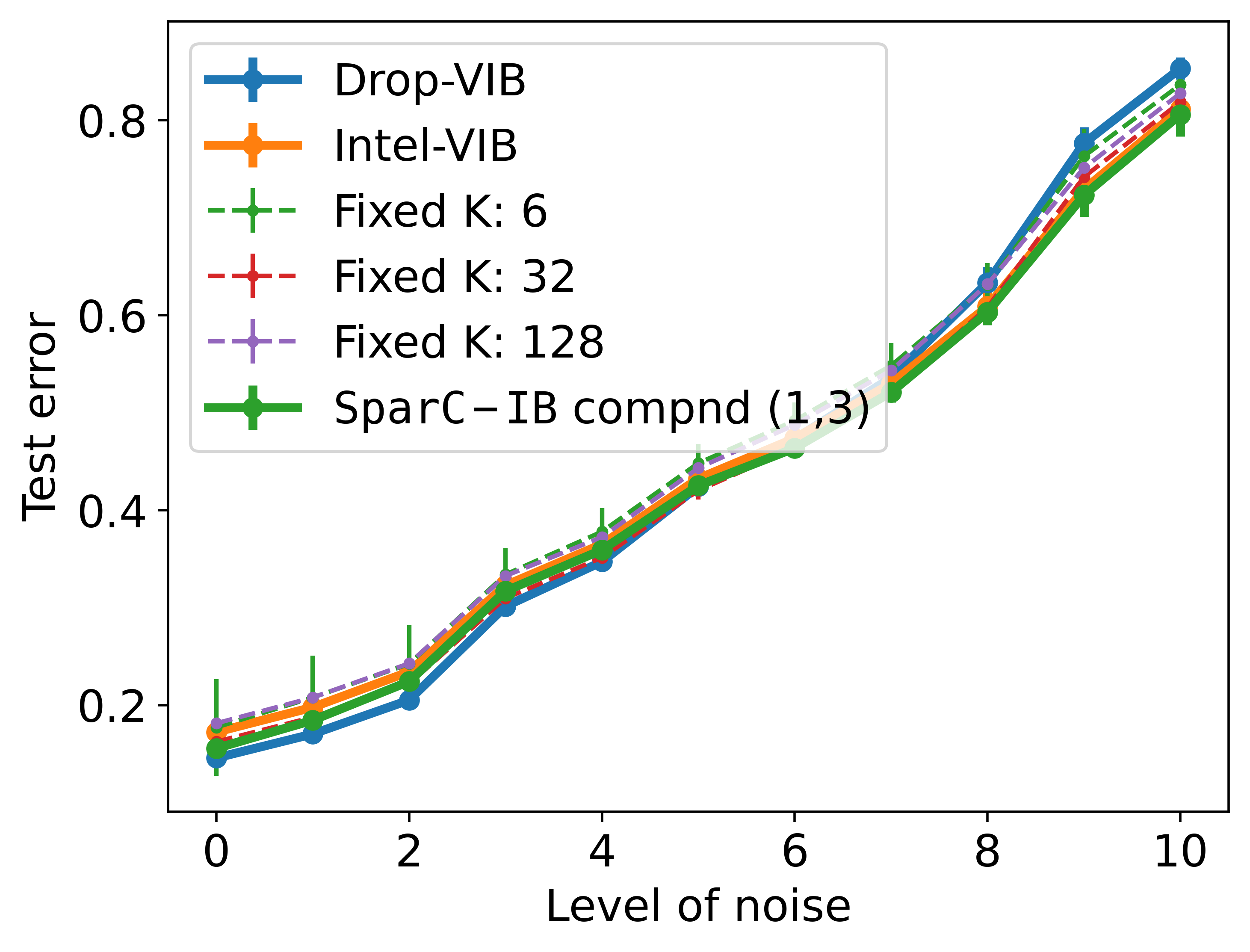}
	\caption{Test error vs noise}
	\end{subfigure}
	\begin{subfigure}{0.3\linewidth}
	\centering
	\includegraphics[width=\linewidth]{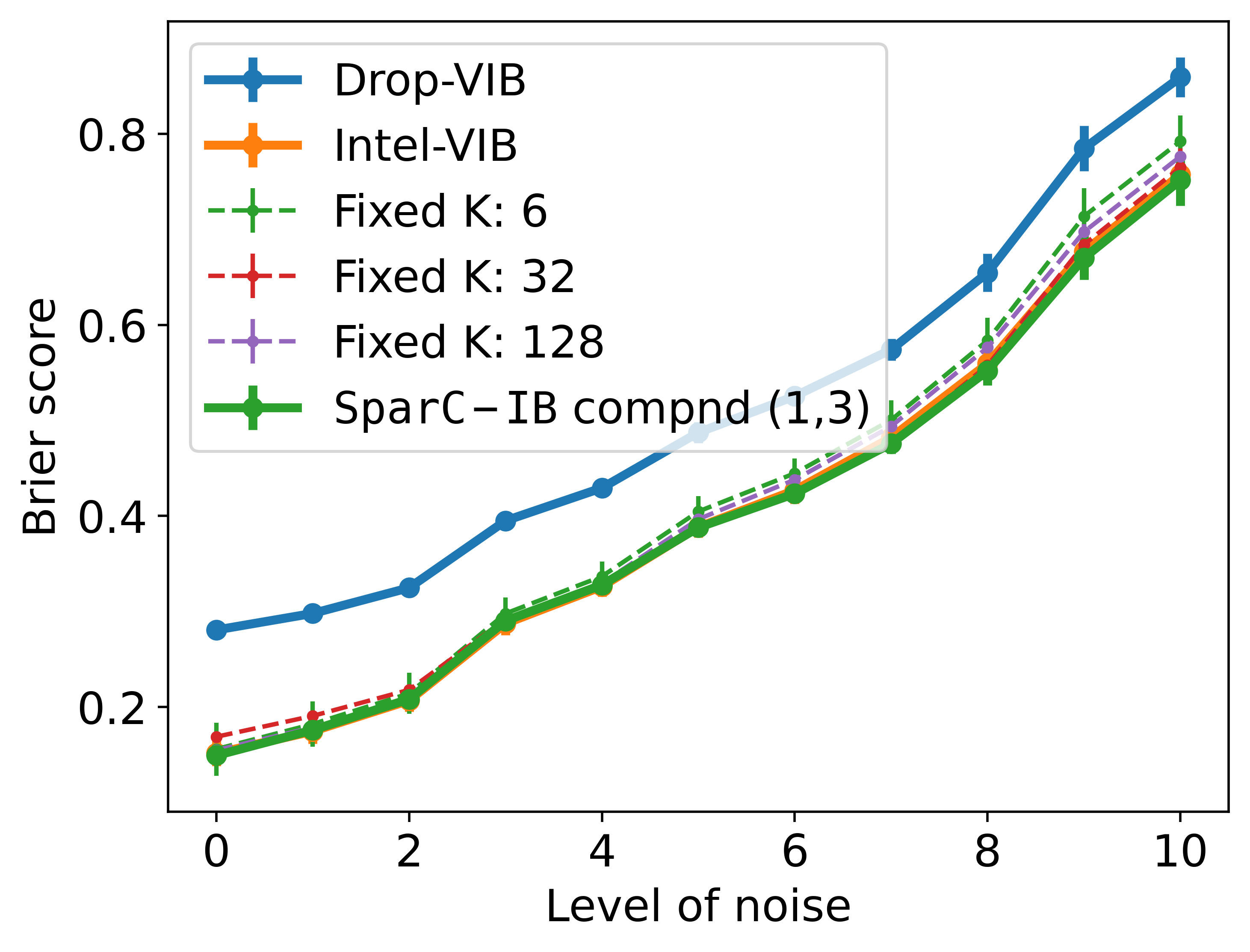}
	\caption{Brier core vs noise}
	\label{brier_v_noise_fmnist}
	\end{subfigure}
	\begin{subfigure}{0.3\linewidth}
	\centering
    \includegraphics[width = \linewidth]{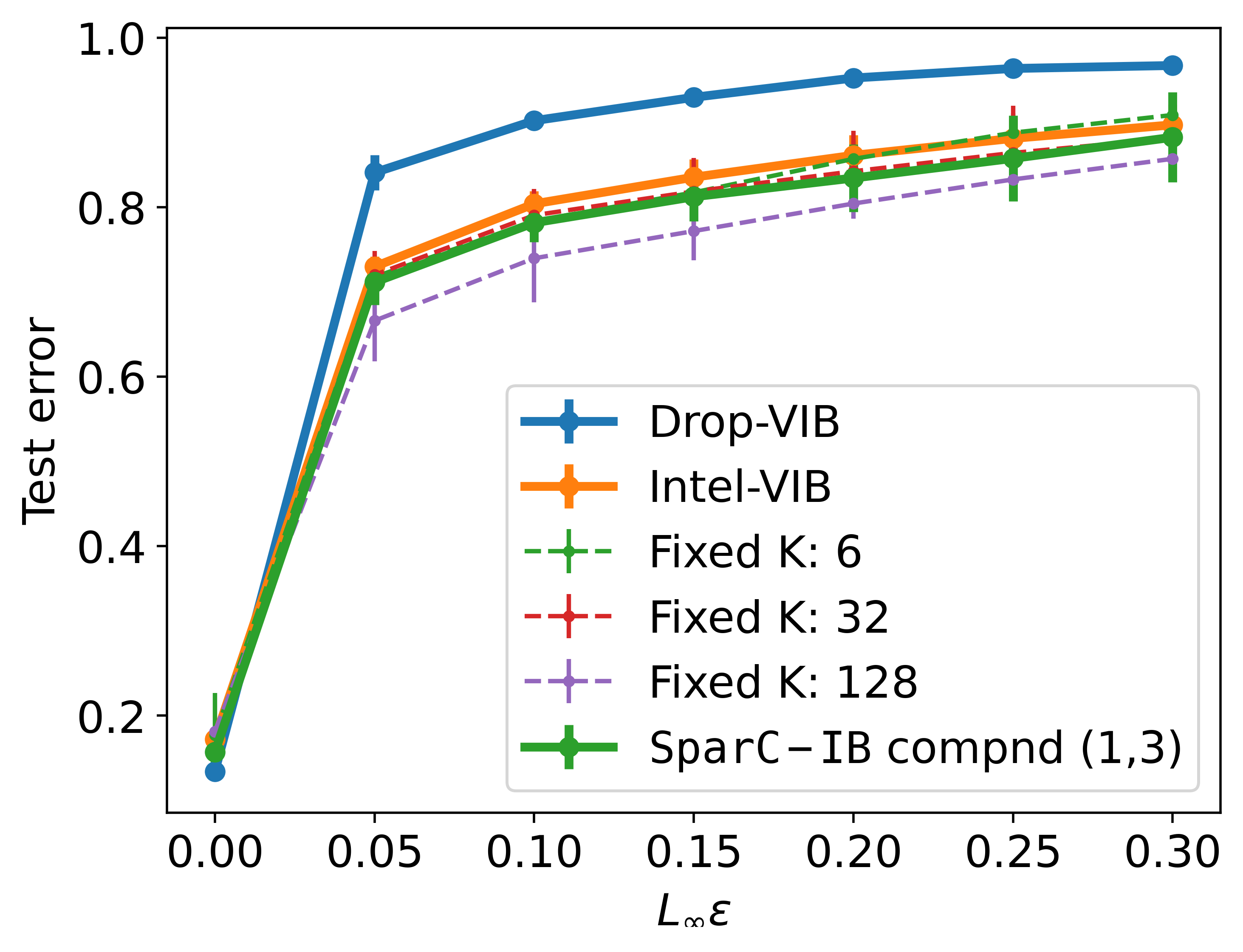}
	\caption{Test error vs $L_\infty$ radius}
	\label{brier_v_noise_fmnist}
	\end{subfigure}
	\caption{Out-of-distribution performance in terms of the test error, and the Brier score on CIFAR-10. We observe that \texttt{SparC-IB} approach with compound strategy and $(a,b)=(1,3)$ (green line) performs as good as the best performing model in all the cases.}
	\label{OOD_noise_fmnist}
\end{figure}

\begin{figure}
	\centering
	\begin{subfigure}{0.3\linewidth}
	\centering
	\includegraphics[width=\linewidth]{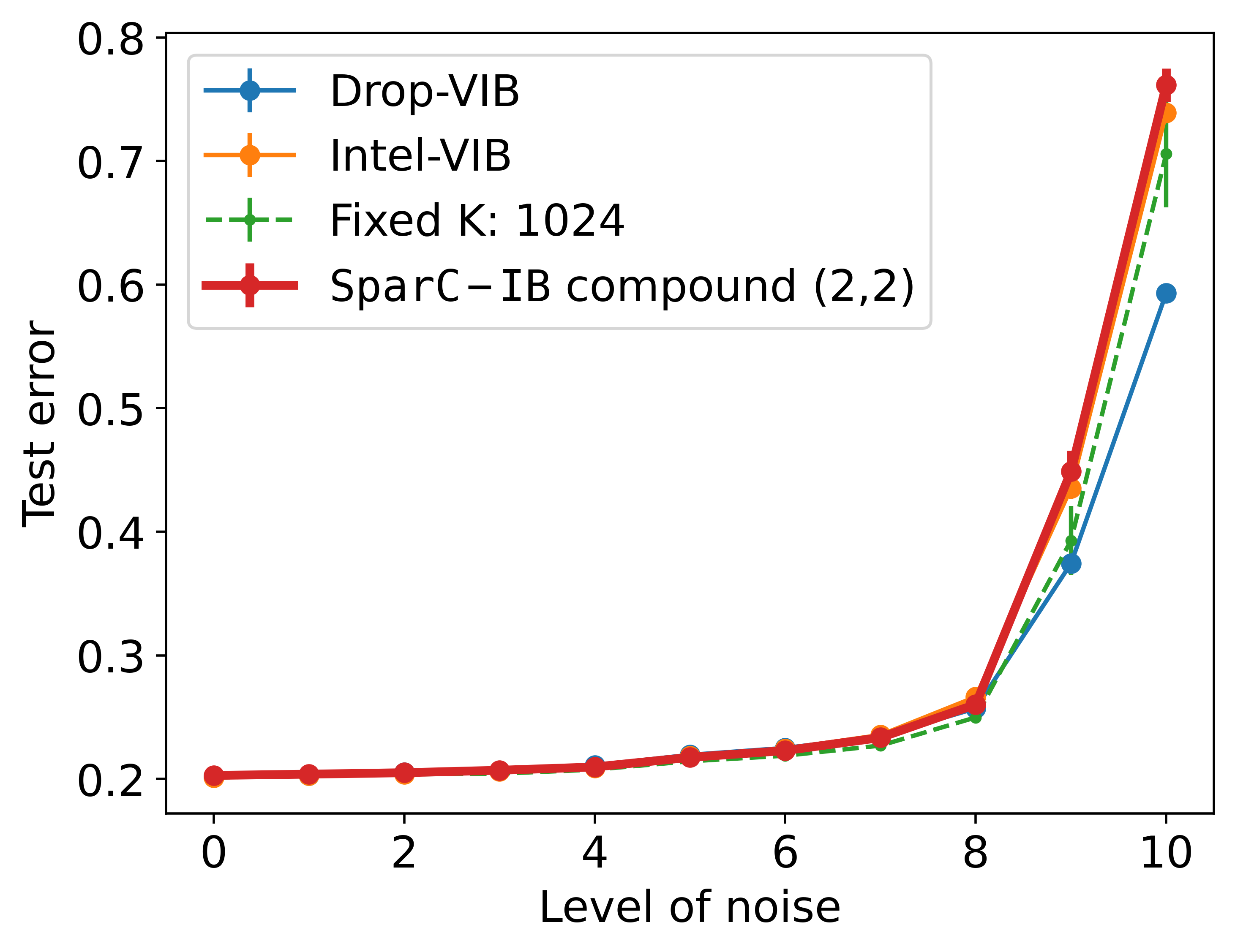}
	\caption{Test error vs noise}
	\end{subfigure}
	\begin{subfigure}{0.3\linewidth}
	\centering
	\includegraphics[width=\linewidth]{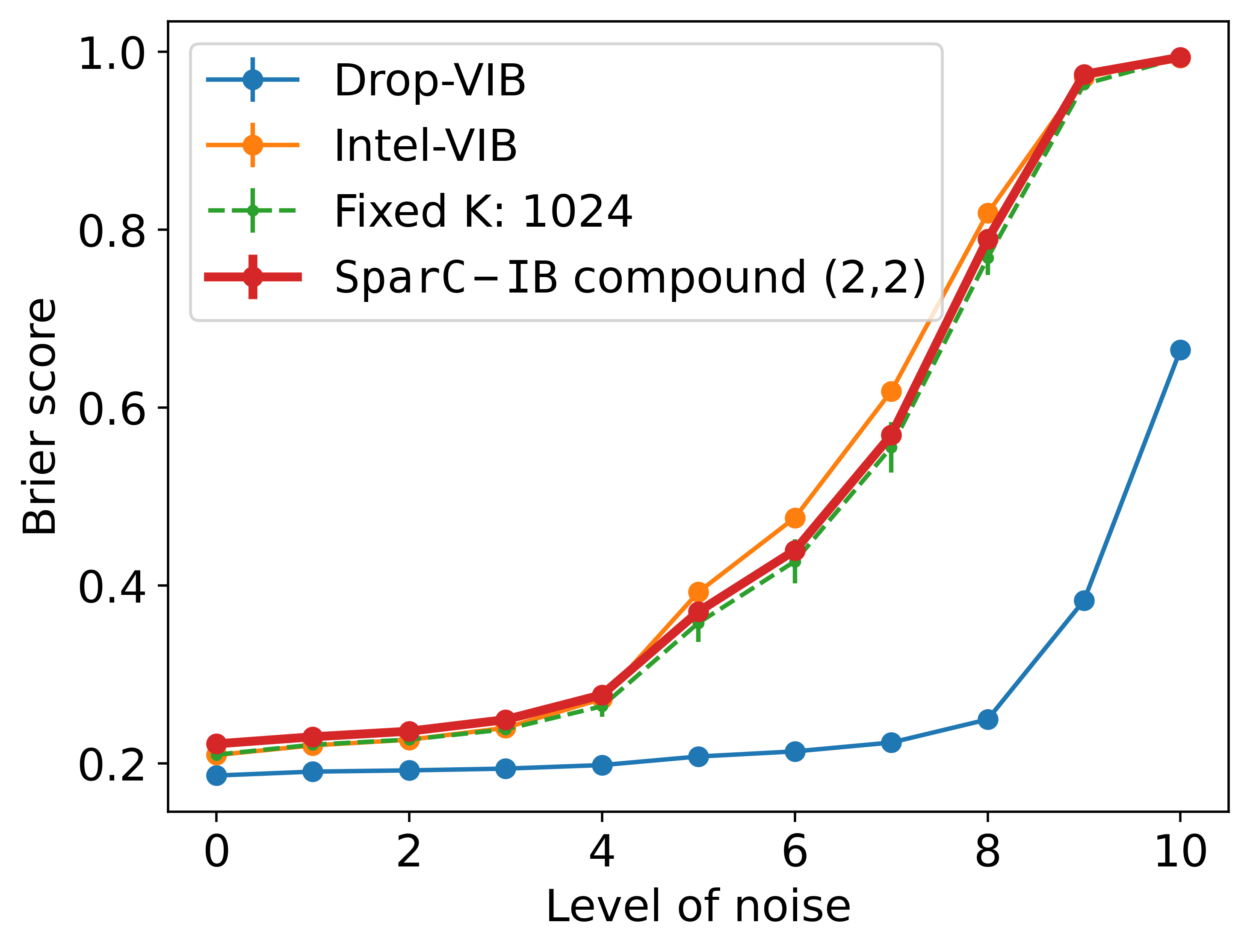}
	\caption{Brier score vs noise}
	\end{subfigure}
	\begin{subfigure}{0.3\linewidth}
	\centering
	\includegraphics[width=\linewidth]{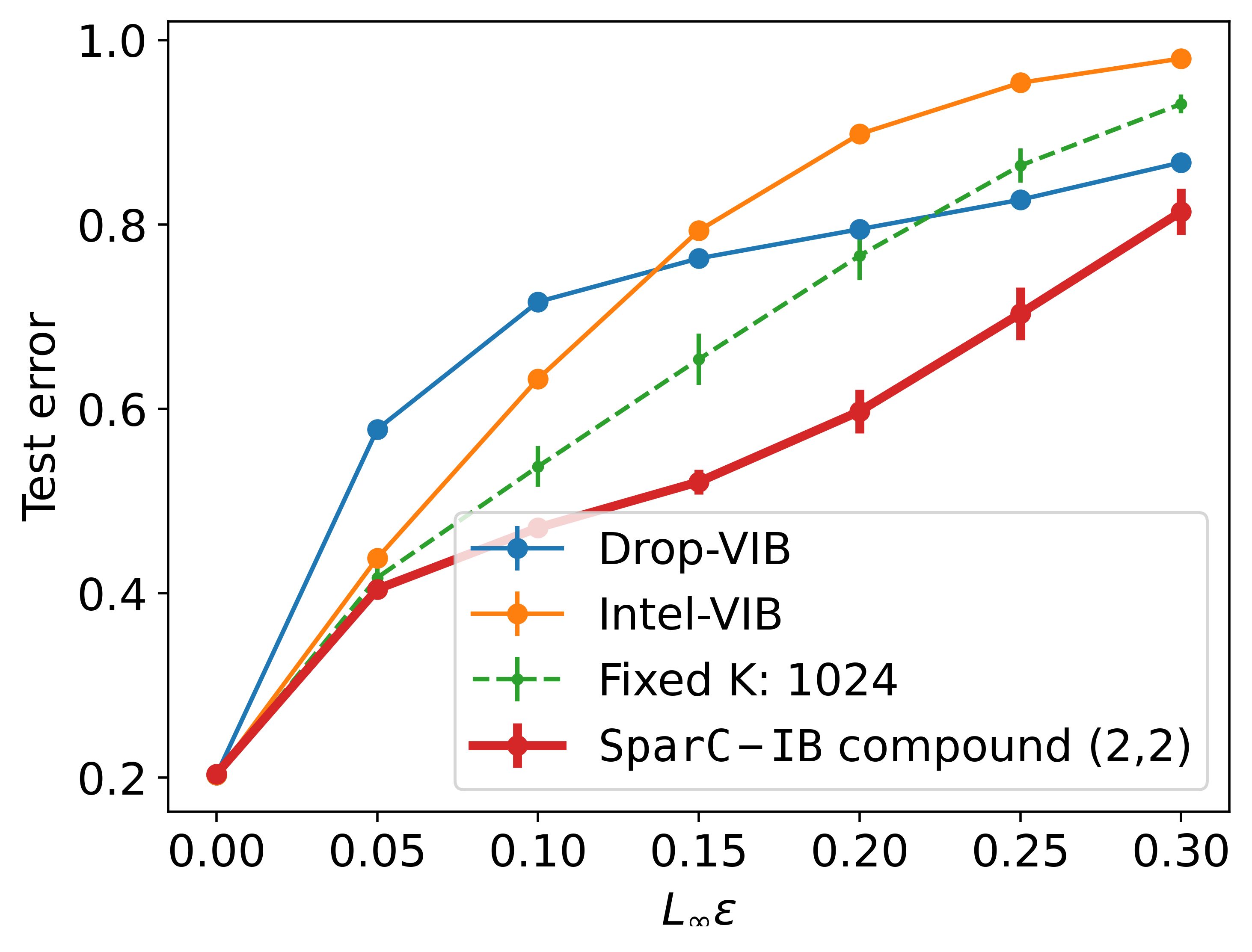}
	\caption{Test error vs $L_\infty$ radius}
	\end{subfigure}
	\caption{Out-of-distribution performance in terms of the test error, and the Brier score on ImageNet. In the white-noise scenario, the drop-VIB performs better than our approach possibly due to learning more information about $X$. However, we observe that \texttt{SparC-IB} approach \emph{outperforms other models in black-box attacks}.}
	\label{fig:BB_ImageNet}
\end{figure}

\subsection{Analysis of the Latent space}
\label{sec::latentspace}
\subsubsection{Dimension distribution mode plot across seeds}
\begin{figure}[h!]
	\centering
	\begin{subfigure}{0.45\linewidth}
	\centering
	\includegraphics[width=\linewidth]{mnist_edit/pow_1-0_repl_1_a_2.0_b_2.0_compound_train_8_no_dim_encoder_Mode_prob_v_digits_joyplot.png}
	\caption{}
	\end{subfigure}
	\begin{subfigure}{0.45\linewidth}
	\centering
	\includegraphics[width=\linewidth]{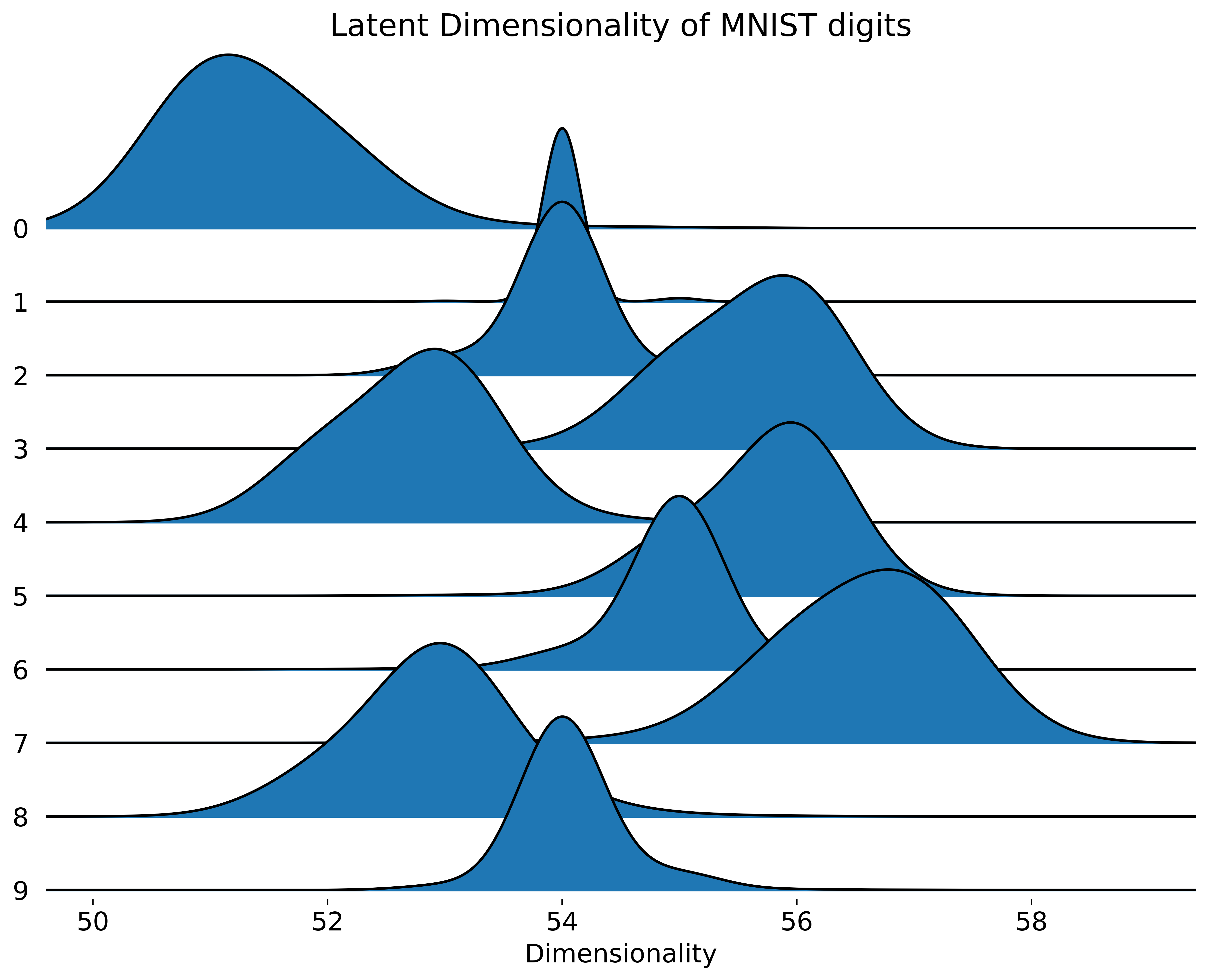}
	\caption{}
	\end{subfigure}
	\begin{subfigure}{0.45\linewidth}
	\centering
	\includegraphics[width=\linewidth]{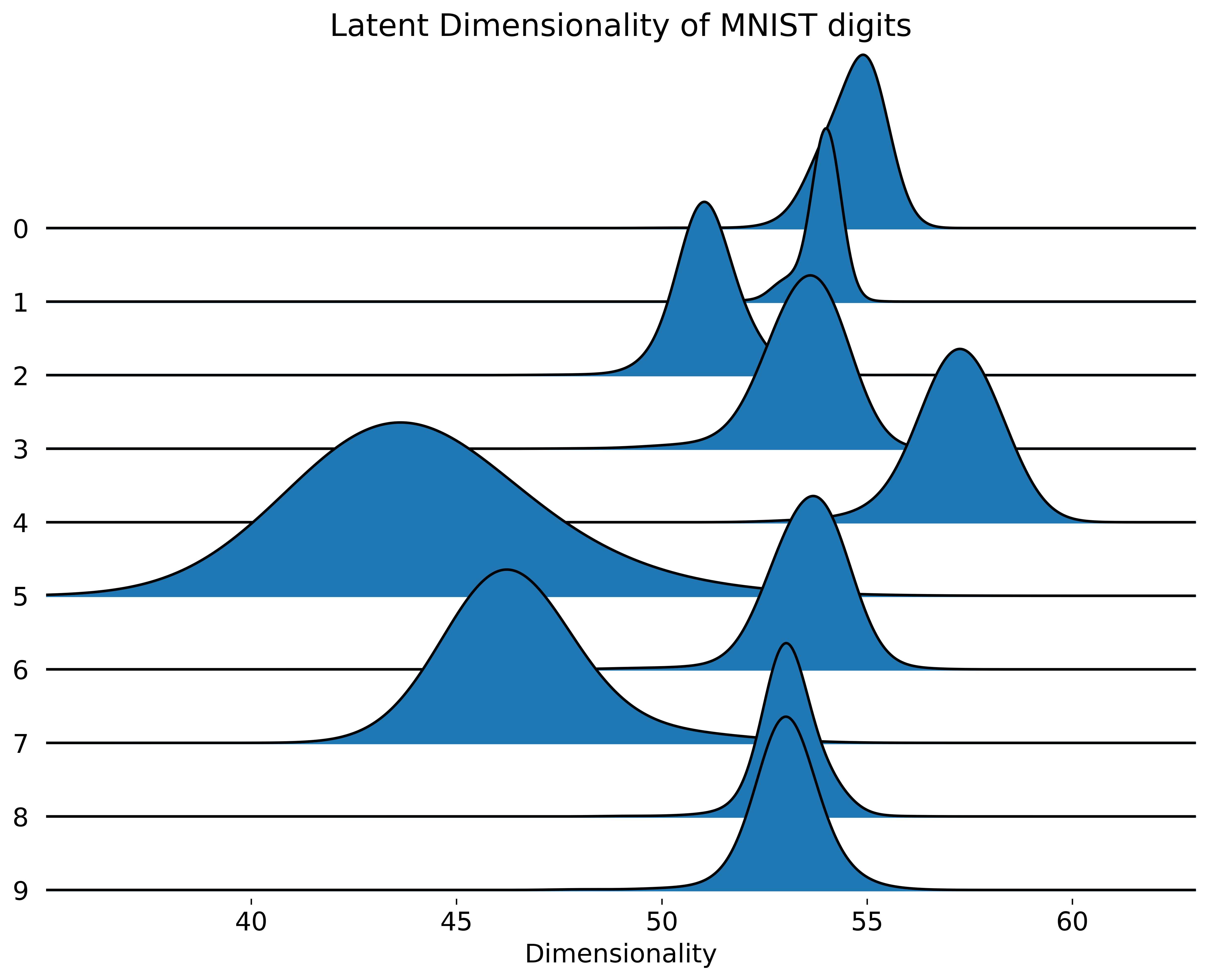}
	\caption{}
	\end{subfigure}
	\caption{Mode plot for \texttt{SparC-IB} compound $(a,b)=(2,2)$ across the 3 seeds on MNIST. We observe separation of posterior modes between the MNIST digits for all 3 seeds.}
	\label{ModePlotSeeds}
\end{figure}

\hyperref[ModePlotSeeds]{Fig. A6} shows the mode plot for \texttt{SparC-IB} compound (2,2) across 3 seeds on MNIST data. The overall range of dimensions remains unchanged across the 3 seeds; however, we observe that each digit prefers a different dimension of the latent space.   

\subsubsection{The dimension distribution mode plot for CIFAR-10 and ImageNet}
\begin{figure}[h!]
	\centering
	\begin{subfigure}{0.45\linewidth}
	\centering
	\includegraphics[width=\linewidth]{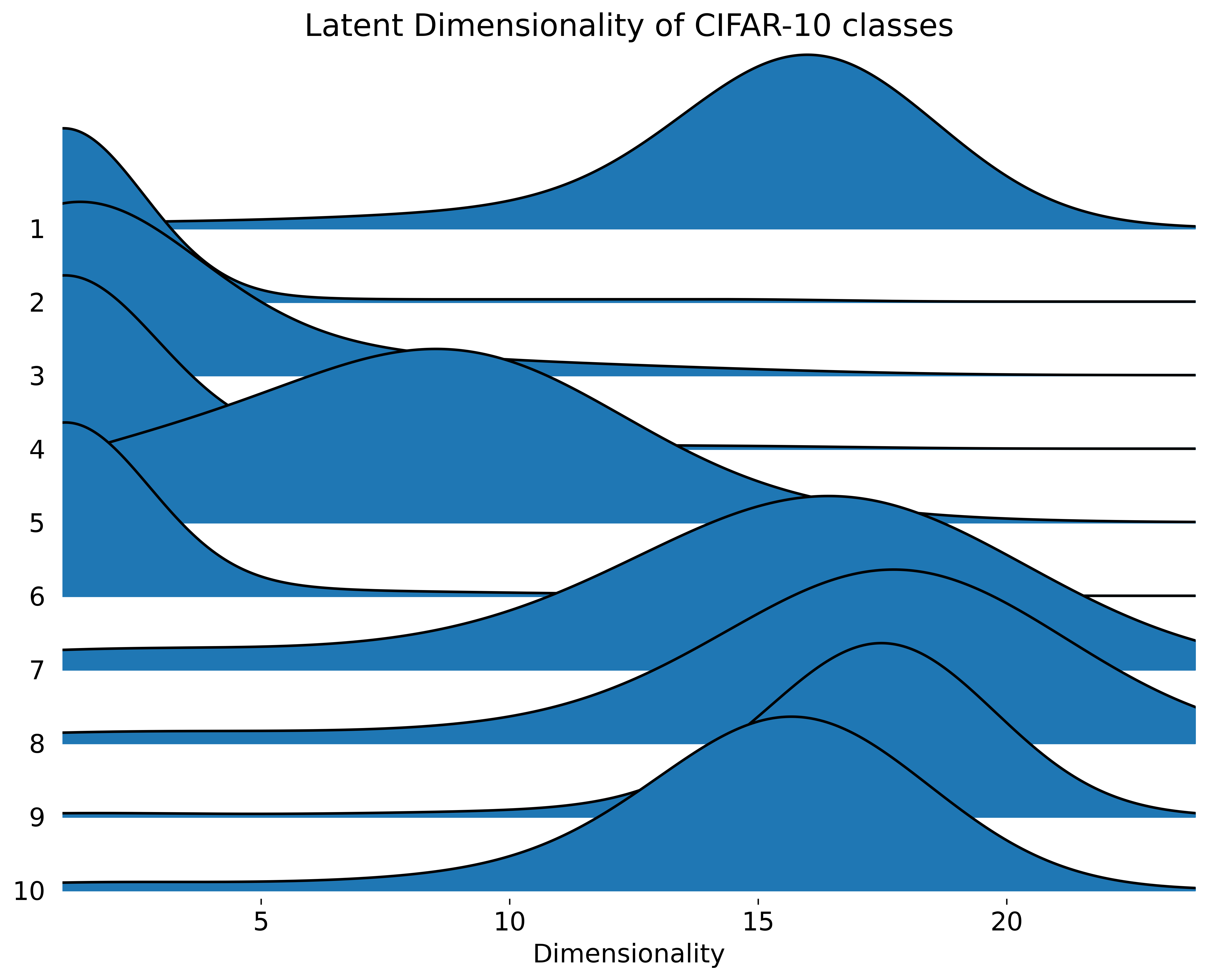}
	\caption{Dimension mode plot for CIFAR-10}
	\label{Modeplot_fmnist}
	\end{subfigure}
	\begin{subfigure}{0.45\linewidth}
	\centering
	\includegraphics[width=\linewidth]{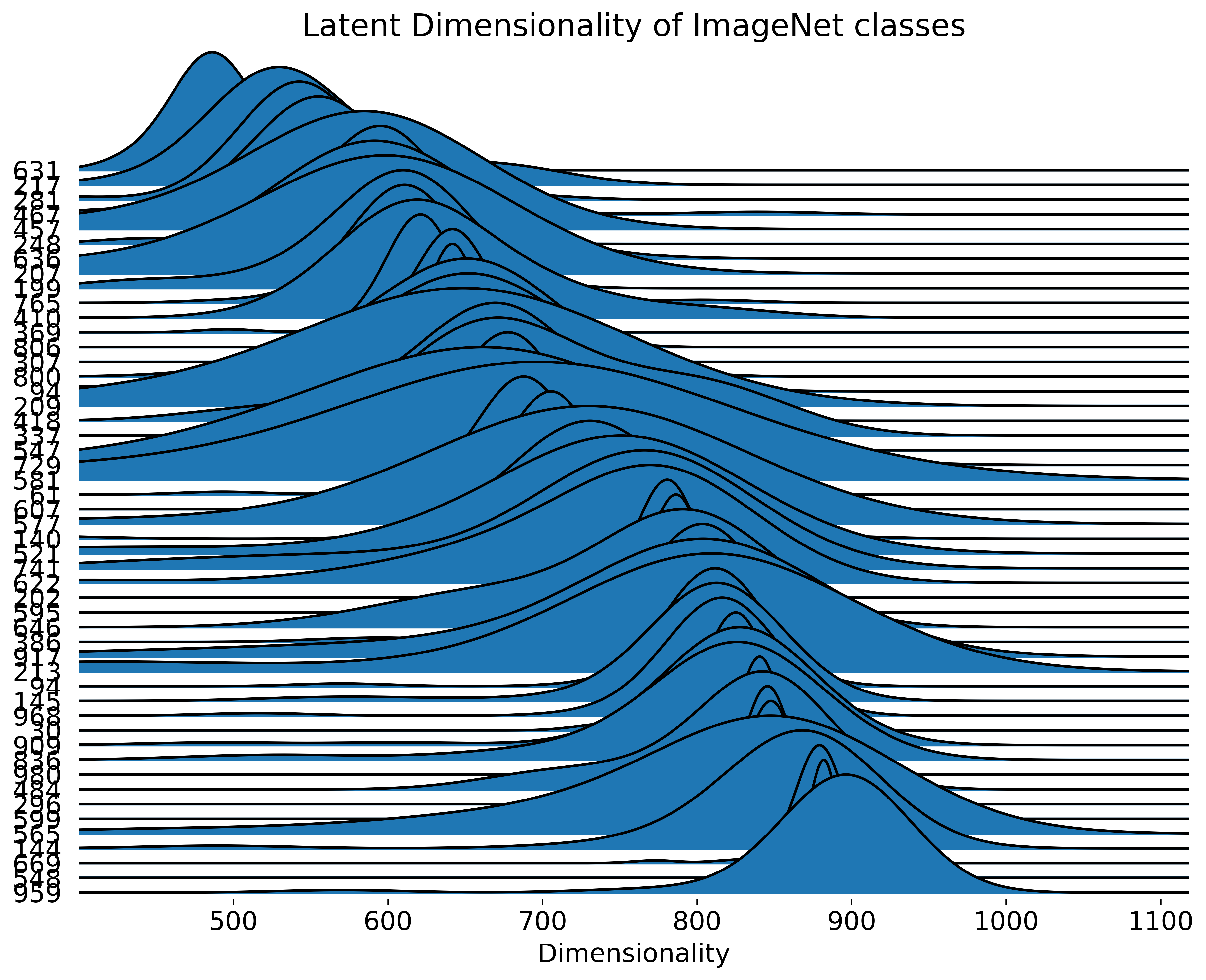}
	\caption{Dimension mode plot for ImageNet.}
	\label{brier_v_beta_othermodels_fmnist}
	\end{subfigure}
	\caption{Plot of the posterior modes of the dimension variable for (a) CIFAR-10 and (b) 50 randomly chosen classes of ImageNet. Both plot show separation between the latent dimension of the classes chosen by \texttt{SparC-IB}.}
	\label{ModePlotCIFARandIMNET}
\end{figure}

\hyperref[ModePlotCIFARandIMNET]{Fig. A7} shows the dimension distribution mode plot across classes of CIFAR-10 and ImageNet. We show these plots for the \texttt{SparC-IB} compound (1,3) in CIFAR-10 and the \texttt{SparC-IB} compound (2,2) in ImageNet. In both data sets, we observe that each class prefers a different latent dimension (especially on ImageNet).

\subsubsection{Information content plot for CIFAR-10 and ImageNet}
\begin{figure}[h!]
	\centering
	\begin{subfigure}{0.45\linewidth}
	\centering
	\includegraphics[width=\linewidth]{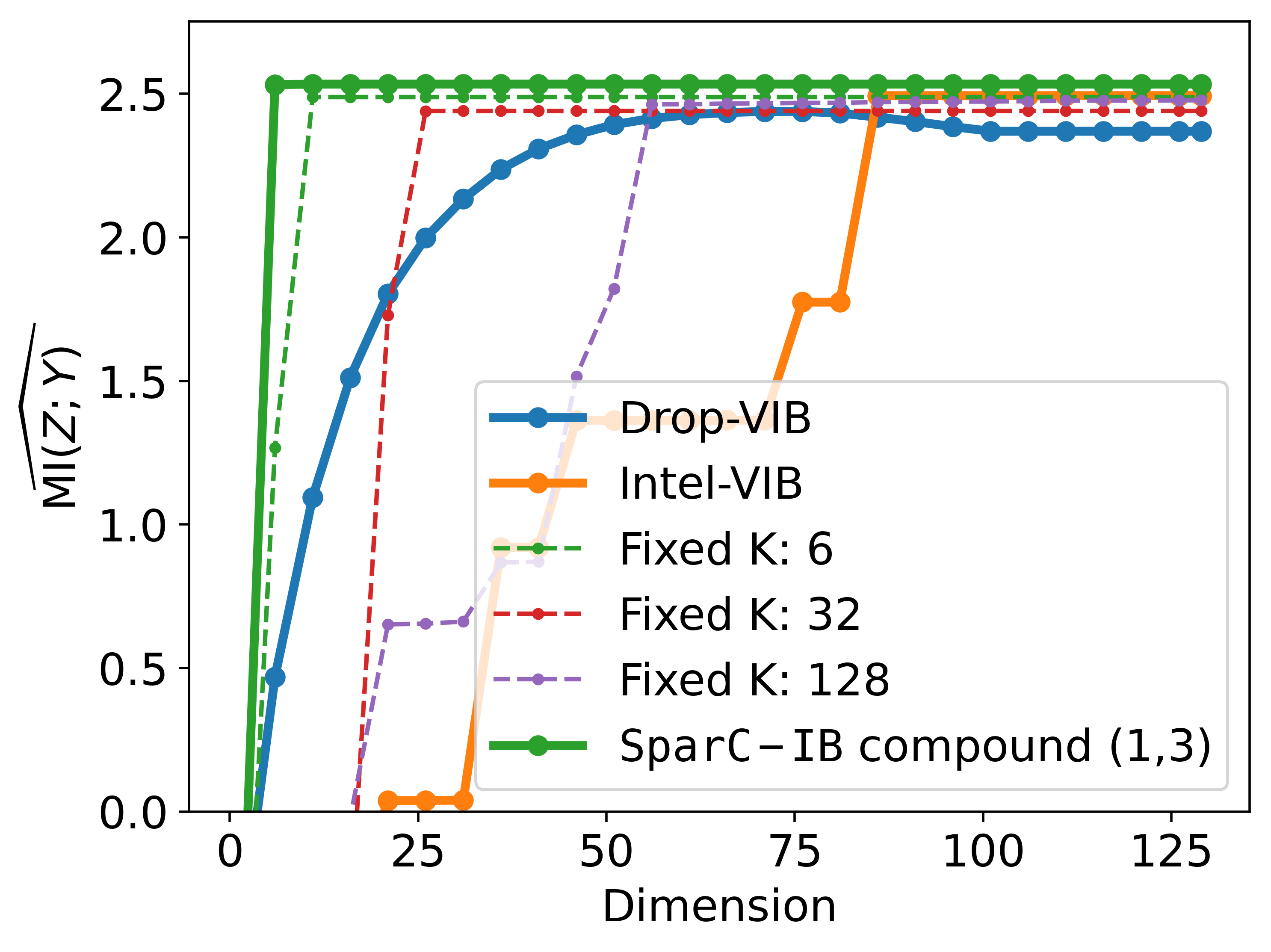}
	\caption{Information content vs dimension plot for CIFAR-10}
	\label{Modeplot_fmnist}
	\end{subfigure}
	\begin{subfigure}{0.45\linewidth}
	\centering
	\includegraphics[width=\linewidth]{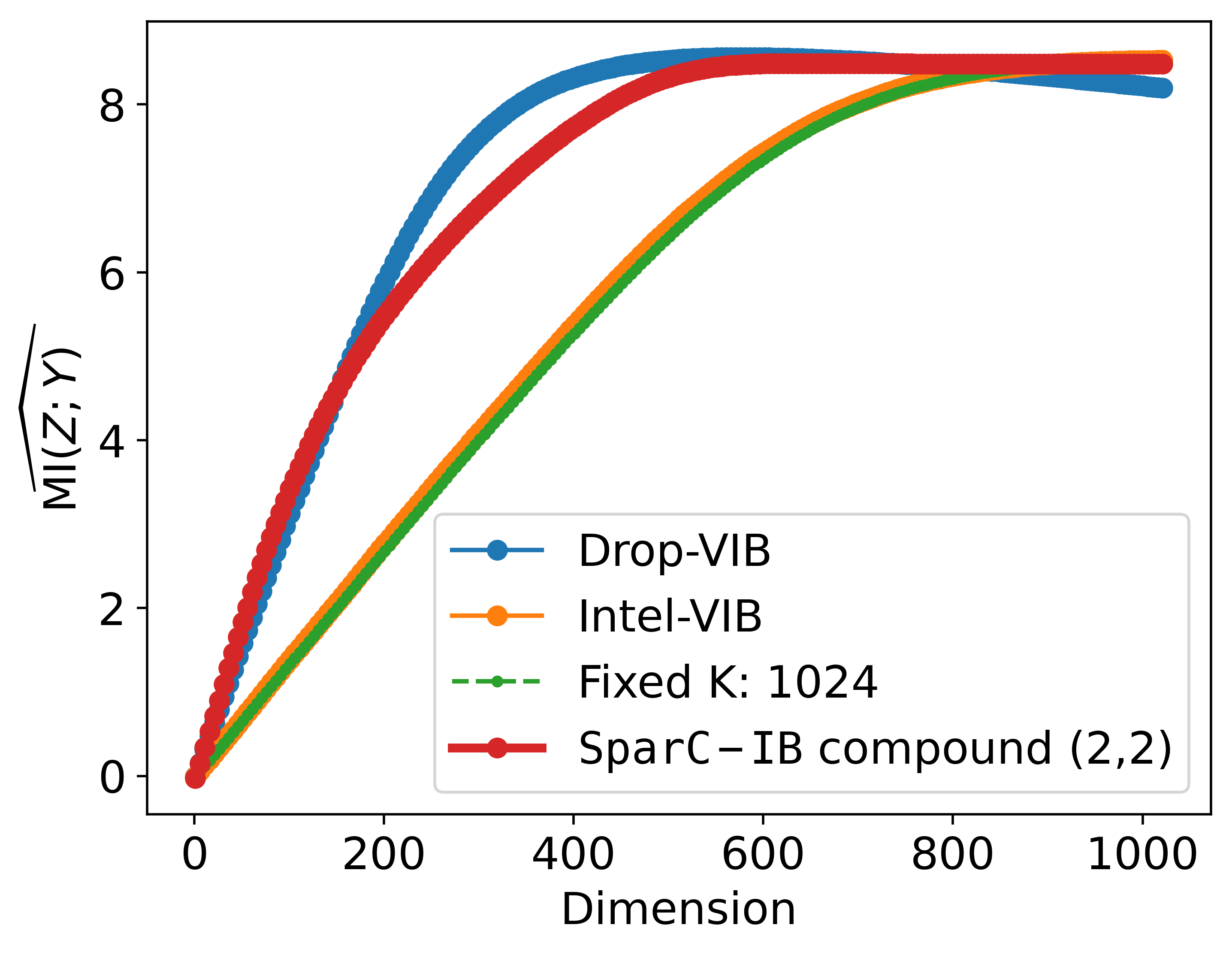}
	\caption{Information content vs dimension plot for ImageNet.}
	\label{brier_v_beta_othermodels_fmnist}
	\end{subfigure}
	\caption{Plot of the information content vs dimension of the mean of the encoder for (a) CIFAR-10 and (b) ImageNet. Both plot show \texttt{SparC-IB} encodes the maximum information in a smaller dimensional latent space than other models.}
	\label{InfoContentCIFARandIMNet}
\end{figure}

\hyperref[InfoContentCIFARandIMNet]{Fig. A8} shows the estimated $\operatorname{MI}(Z;Y)$ (expression provided in \hyperref[sec::InfoCont]{Sec. 4.3.2}) against the increasing dimension of the latent space for CIFAR-10 and ImageNet. In CIFAR-10, \texttt{SparC-IB} provides the most compact representation among the other models in which the information plateaus within the first dimensions ($\sim$ 5) of the latent space. For ImageNet, we observe that the information plateaus around dimension 500 which is smaller than the fixed-dimensional VIB and Intel-VIB models but higher than the Drop-VIB model. In addition, we note that the behavior of the estimated $\operatorname{MI}(Z;Y)$ as a function of dimension is much smoother than those of the other two data sets. The reason for such behavior is perhaps the complexity in the ImageNet data, where it requires a high-dimensional latent space to encode the necessary information of $X$ where each dimension's contribution is small.  

\subsubsection{Calculating pixel-wise importance scores for latent dimension visualization on MNIST}
\label{sec::calc_ImpScore}
We have used the Captum package \cite{kokhlikyan2020captum} to calculate the pixel importance scores to visualize the latent space (\hyperref[sec::VizLatent]{Sec. 4.3.3}) for MNIST. Given an input image $x$ and a baseline image $x'$, the importance score for the $i$ th pixel on the $d$ th dimension of the mean of the latent space is calculated using the following expression.
\begin{flalign}
\text{Importance Score}_i^d (x) &= (x_i - x_i') \int_{\alpha=0}^{1} \frac{\partial \mu_d(\alpha x + (1-\alpha) x')}{\partial x_i} d\alpha & \nonumber
\end{flalign}
In the above expression, $\mu(.)$ is the mean vector of the latent space and $\mu_d(.)$ represents its $d$-th coordinate. We used a blank image where every pixel value is 0 as a baseline $x'$. We can interpret the score as the sensitivity of the dimensions of $\mu(x)$ to a small change in each pixel integrated on the images that fall on the line given by $\alpha x + (1-\alpha) x'$.

\end{document}